\documentclass[lettersize,journal]{IEEEtran}
\usepackage{amsmath,amsfonts}
\usepackage{algorithmic}
\usepackage{algorithm}
\usepackage{array}
\usepackage[caption=false,font=normalsize,labelfont=sf,textfont=sf]{subfig}
\usepackage{textcomp}
\usepackage{stfloats}
\usepackage{url}
\usepackage{verbatim}
\usepackage{graphicx}
\usepackage{cite}
\usepackage{colortbl}
\usepackage{amsfonts}
\usepackage{multirow}
\usepackage{multicol}
\usepackage{booktabs}
\usepackage{xcolor}
\usepackage{makecell}
\usepackage{float}

\begin{document}

\title{Simulate, Refocus and Ensemble: An Attention-Refocusing Scheme for Domain Generalization}

\author{Ziyi Wang, Zhi Gao, Jin Chen, Qingjie Zhao, Xinxiao Wu$^{*}$,~\IEEEmembership{Member,~IEEE,} Jiebo Luo,~\IEEEmembership{Fellow,~IEEE}
\thanks{Ziyi Wang, Zhi Gao, Jin Chen, Qingjie Zhao are with the Beijing Key Laboratory of Intelligent Information Technology, School of Computer Science and Technology, Beijing Institute of Technology, Beijing
        100081, China.}
\thanks{Xinxiao Wu is  with the Guangdong Laboratory of Machine Perception and Intelligent Computing, Shenzhen MSU-BIT University, Shenzhen 518172, and also with  the Beijing Key Laboratory of Intelligent Information Technology, School of Computer Science and Technology, Beijing Institute of Technology, Beijing 100081, China.}% <-this % stops a space
% \thanks{Zhi Gao is with the School of Intelligence Science and Technology, Peking University, also with the Beijing Institute for General Artificial Intelligence, Beijing, China (e-mail: zhi.gao@pku.edu.cn).}% <-this % stops a space
% \thanks{Jin Chen is with the Intelligent Science \& Technology Academy of CASIC, Beijing, China (e-mail: chenjinbit@163.com).}% <-this % stops a space
\thanks{Jiebo Luo is with the Department of Computer Science, University of Rochester, Rochester, NY 14627 USA.}
\thanks{*Corresponding author: Xinxiao Wu (e-mail: wuxinxiao@bit.edu.cn).}
}

% \author{IEEE Publication Technology,~\IEEEmembership{Staff,~IEEE,}
%         % <-this % stops a space
% \thanks{This paper was produced by the IEEE Publication Technology Group. They are in Piscataway, NJ.}% <-this % stops a space
% \thanks{Manuscript received April 19, 2021; revised August 16, 2021.}}

% The paper headers
%\markboth{Journal of \LaTeX\ Class Files,~Vol.~14, No.~8, August~2021}%
%{Shell \MakeLowercase{\textit{et al.}}: A Sample Article Using IEEEtran.cls for IEEE Journals}

% \IEEEpubid{0000--0000/00\$00.00~\copyright~2021 IEEE}
% Remember, if you use this you must call \IEEEpubidadjcol in the second
% column for its text to clear the IEEEpubid mark.

\maketitle

\begin{abstract}
Domain generalization (DG) aims to learn a model from source domains and apply it to unseen target domains with out-of-distribution data. Owing to CLIP's strong ability to encode semantic concepts, it has attracted increasing interest in domain generalization.
However, CLIP often struggles to focus on task-relevant regions across domains, \emph{i.e.,} domain-invariant regions, resulting in suboptimal performance on unseen target domains.
To address this challenge, we propose an attention-refocusing scheme, called \emph{Simulate, Refocus and Ensemble (SRE)}, which learns to reduce the domain shift by aligning the attention maps in CLIP via attention refocusing.
SRE first simulates domain shifts by performing augmentation on the source data to generate simulated target domains. SRE then
 learns to reduce the domain shifts by refocusing the attention in CLIP between the source and simulated target domains.
% \jin{Finally, SRE applys an ensembling learning manner to integrate parameters that are robust to attention refocusing for enhancing the learning of it.}
%Then it reduces the simulated domain shifts by employing an attention refocuser to align attention maps in CLIP of the source data and the augmented views. 
%Finally, SRE utilizes ensemble learning to select and combine parameters of the attention refocuser, enhancing to capture domain-invariant attention maps between the source data and the simulated target data.
Finally, SRE utilizes ensemble learning to enhance the ability to capture domain-invariant attention maps between the source data and the simulated target data.
Extensive experimental results on several datasets demonstrate that SRE generally achieves better results than state-of-the-art methods. The code is available at: \url{https://github.com/bitPrincy/SRE-DG}.
%show the effectiveness of  SRE. Our method achieves 2.1\% performance gain on the TerraIncognita dataset compared to the previous SOTA approach.
\end{abstract}

\begin{IEEEkeywords}
Domain generalization; CLIP; Attention refocusing
\end{IEEEkeywords}

\section{Introduction}
\IEEEPARstart{D}{omain} Generalization (DG) aims to train models that can perform well on unseen target domains of different distributions by using labeled data from source domains~\cite{chen2023meta,lv2022causality,shankar2018generalizing,mahajan2021domain,zhou2020learning,nam2021reducing}.
%Recently, CLIP~\cite{radford2021learning} has emerged as a promising candidate for domain generalization~\cite{10377071,li2022learning,niu2022domain}, since it is pre-trained on large-scale data and encodes rich semantic concepts, showing better generalization ability than traditional neural networks~\cite{radford2021learning,10.5555/3666122.3666725,zhao2024testtime}.
\textcolor{black!70!black}{Recently, CLIP~\cite{radford2021learning} has emerged as a promising candidate for domain generalization~\cite{10377071,li2022learning,niu2022domain}, since it is pre-trained on larger scale data and encodes richer semantic concepts than traditional neural networks~\cite{radford2021learning,10.5555/3666122.3666725,zhao2024testtime,cheng2025achieving,cheng2024progressive,cheng2024continual}.}

\begin{figure}[tb]
	\centering
		\includegraphics[width=0.46\textwidth]{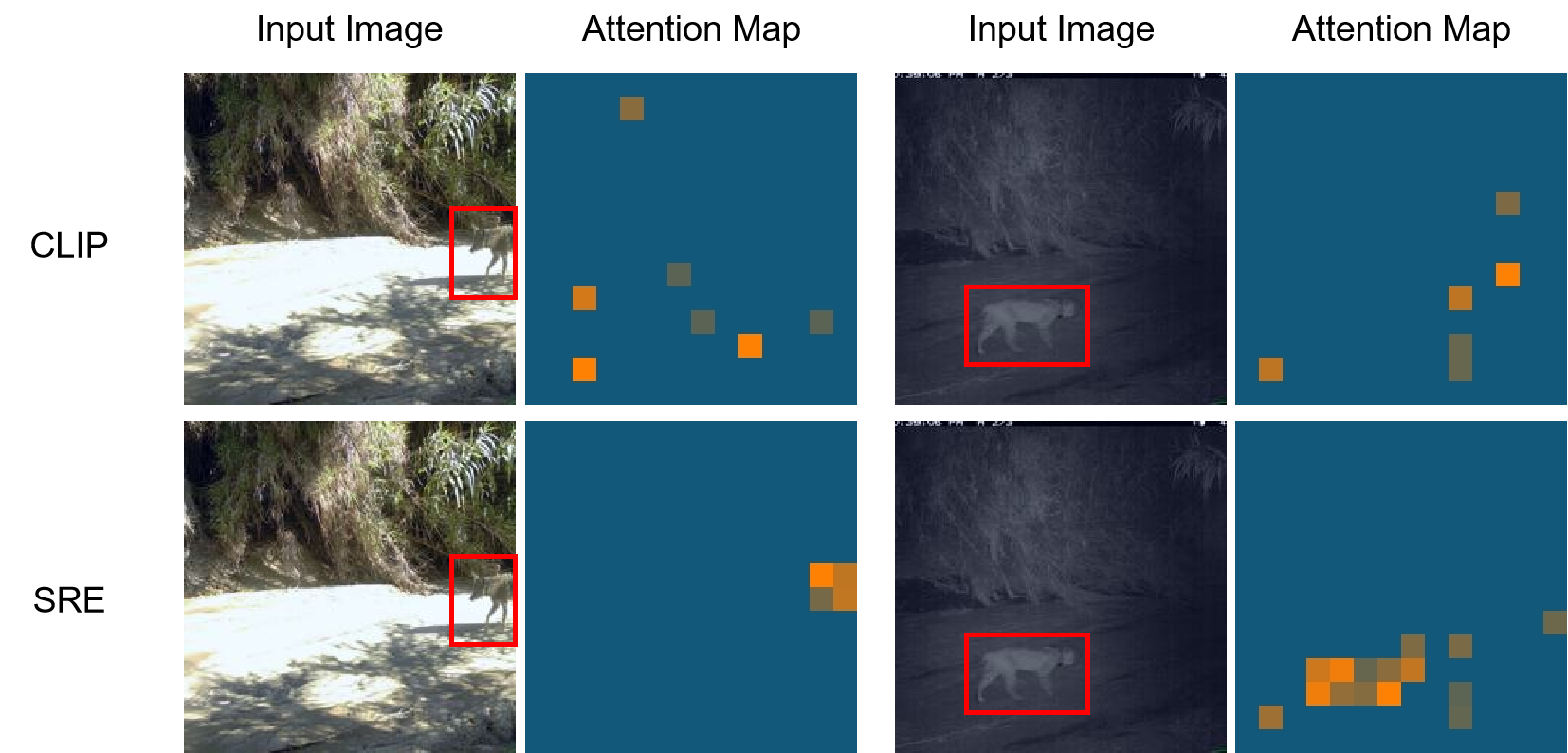}
	% \subfigure[Original and refined attention maps for the task-level distribution shifts]{
		% \includegraphics[width=0.45\textwidth]{img/refine.png}}
	
	\caption{Attention maps of input images generated by CLIP and the proposed SRE. CLIP often cannot focus on task-relevant regions (\emph{i.e.}, regions of the animals), while SRE can focus on those regions, thus facilitating the subsequent classification.
		% (c) CLIP tends to focus on not only object-related regions of the image but also some object-unrelated regions, due to the task-level distribution shifts.
	}
	\label{fig:intro2}
\end{figure}

However, CLIP has not yet achieved excellent performance in domain generalization. One important reason is that CLIP tends to understand images from a global perspective, as it is trained by aligning images and texts instead of learning discriminative features for classification.
% , \emph{i.e.,} it struggles to always focus on task-relevant regions.
Hence, CLIP often struggles to focus on task-relevant regions, making it difficult to capture domain-invariant representations.
% losing domain-invariant information and limiting the performance in domain generalization.
As shown in the top part of Figure~\ref{fig:intro2}, the attention maps of input images generated by CLIP mainly focus on the background regions unrelated to classification. %while our method achieve better focus on the animals. 
% As CLIP fails to focus on the task-relevant regions, the domain-invariant representations are hard to capture, resulting in inferior performances on the task of domain generalization.
%The failure of CLIP  focus on task-relevant regions results in the absence of domain-invariant representations and inferior domain generalization performance.

To address the above challenge, we introduce attention refocusing into CLIP, thus reducing the domain shift by refocusing the attention across domains.
%which adapts the attention mechanism for unseen target domains to better capture domain-invariant representations.
Attention refocusing usually uses additional attention masks~\cite{sun2024alpha}, fine-tuned attention weights~\cite{phung2024grounded}, or a learned bias to modify the attention mechanism of inputs~\cite{shi2023toast}.
% Thanks to its ability to direct 
In this case, the model's attention is directed to task-relevant image regions, showing effectiveness in multiple visual tasks, such as image synthesis~\cite{wang2024primecomposer, phung2024grounded} and image classification~\cite{yoo2021xecgnet, shi2023toast}.

% \textcolor{black}{A recent study~\cite{yao2024theoretical} provides a theoretical analysis of fine-tuning attention mechanisms, demonstrating that adjusting specific components of the attention mechanism can improve generalization bounds and enhance memory efficiency. The authors show that fine-tuning the query and value matrices within the attention mechanism leads to better generalization compared to fine-tuning all components, offering insights into the optimization dynamics of attention-based models.}

% \textcolor{black}{In domain generalization, the objective is to develop models that perform effectively on unseen target domains by leveraging knowledge from multiple source domains. Attention mechanisms contribute to this goal by emphasizing domain-invariant features while minimizing reliance on domain-specific attributes~\cite{meng2022attention}.}

Inspired by this, we introduce an attention-refocusing scheme for domain generalization, by which CLIP adaptively focuses more on task-relevant regions containing domain-invariant information, bridging the source and unseen target domains.

\begin{figure}[tb]
	\centering
		\includegraphics[width=0.45\textwidth]{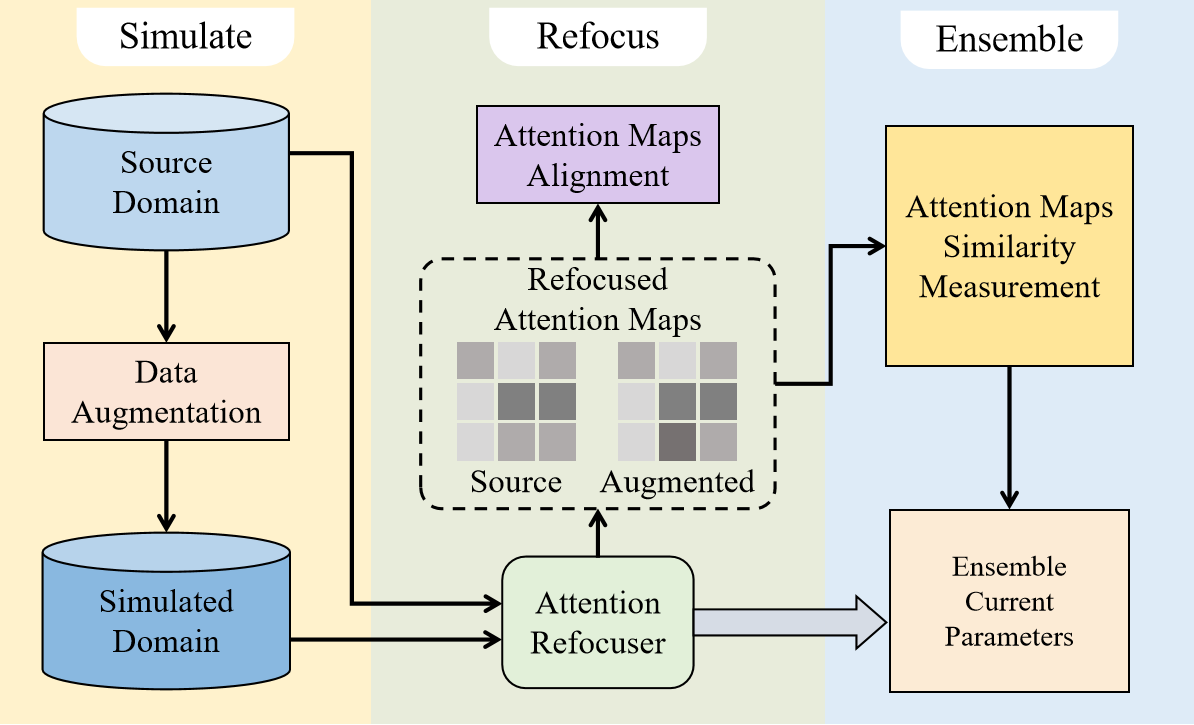}
	\caption{An overview of the proposed DG attention-refocusing scheme SRE. 
	}
	\label{fig:scheme}
\end{figure}

In this paper, we propose a three-stage DG scheme, named \emph{Simulate, Refocus and Ensemble (SRE)}, 
%as displayed in Figure~\ref{fig:scheme}, 
where the domain shifts are simulated and bridged over the attention maps of data. The \emph{Simulate} stage generates simulated target domains by augmenting source images with consistent spatial relationships. 
%Instead of directly fine-tuning the attention weights, 
The \emph{Refocus} stage introduces an attention-refocuser %based on TOAST~\cite{shi2023toast} 
to capture the data distributions of the source images and simulated target images, and align the attention maps of the two kinds of images based on their data distributions.
The \emph{Ensemble} stage selects the parameters of the attention-refocuser with high-consistent attention between the source and the simulated target images, filtering out the parameters that are skewed on challenging images (e.g., %the color and texture are similar between object and background
the object and background are similar in color and texture). Figure~\ref{fig:scheme} shows the overview of SRE.

%Compared with existing methods for reducing domain shifts, 
Our scheme benefits from attention refocusing and allows CLIP to adaptively focus on domain-invariant regions within each sample, thereby learning domain-invariant features. 
Different from existing attention-refocusing methods, such as TOAST~\cite{shi2023toast} that requires an additional training stage to refocus on task-relevant regions in downstream tasks, our scheme learns to perform refocusing on diverse simulated domains, through which it can effectively adapt to unseen target domains without any extra training.
% does not rely on any pre-tuning and achieves much better
% pre-tuning on additional data to achieve meaningful performance gains, our scheme does not rely on any pre-tuning and achieves much better generalization.
% Unlike TOAST, which requires pre-tuning on additional data to achieve meaningful performance gains, our scheme does not rely on any pre-tuning and achieves much better generalization.
Experimental results demonstrate that our method achieves competitive performance compared to state-of-the-art methods, verifying that attention refocusing significantly improves generalization.

%The contributions of this work can be summarized as follows:
%\begin{itemize}
%    \item We propose a novel DG attention-refocusing scheme, \emph{Simulate, Refocus and Ensemble}, which balances good domain generalization with the ability to encode rich semantic concepts in CLIP.
%    \item We propose a novel method under this scheme, where the domain shift between source and target domains is simulated by data augmentation and bridged by aligning attention maps between source and augmented images while robust parameters are selected and ensembled to avoid training instability.
%    \item \jin{Experiments}
%\end{itemize}
% (1) We propose a novel DG attention-refocusing scheme, \emph{Simulate, Refocus and Ensemble}, which balances good domain generalization with the ability to encode rich semantic concepts in CLIP.
% (2) We propose a novel method under this scheme, where the domain shift between source and target domains is simulated by data augmentation and bridged by aligning attention maps between source and augmented images while robust parameters are selected and ensembled to avoid training instability.

% (2) We propose to simulate the domain shift and then learn to reduce it by adaptively adjusting the attention of data in CLIP to focus on domain-invariant task-relevant regions.
% Besides, we propose to select parameters that consistently focus on domain-invariant regions between two augmented images, thereby alleviating training instability caused by challenging data.

The remainder of this paper is organized as follows. In Section II, we summarize previous works related to our method. In Section III, we illustrate the proposed attention-refocusing scheme SRE for domain generalization. Section IV demonstrates experimental results on various benchmark datasets and the conclusion is presented in Section V.

\section{Related Work}
\subsection{Domain Generalization}
Domain generalization (DG) aims to develop models that can generalize well to unseen target domains given data from multiple source domains. This problem is critical in real-world applications where models face significant domain shifts, such as changes in environments, sensor configurations, or image styles. Unlike domain adaptation, which assumes access to unlabeled or labeled data in the target domain during training, DG methods operate under a more rigorous setting where no target domain information is available, making it a more challenging yet more practical research problem.

Traditional domain generalization methods primarily include data augmentation, representation learning, and learning strategy. \textcolor{black!70!black}{The data augmentation methods~\cite{wang2021learning,zhou2020learning,nam2021reducing,qiao2020learning,huang2024representation} diversify the source domain by perturbing original data and expanding the data distribution within the source domain.}
% M-ADA~\cite{qiao2020learning} employs adversarial training to create adversarial simulated target data and uses a Wasserstein Auto-Encoder~\cite{tolstikhin2018wasserstein} to relax the semantic consistency constraint used in previous methods. 
% L2D~\cite{wang2021learning} introduces a style-complement module that minimizes the upper bound of mutual information between the original and synthesized images, thereby diversifying the generated images.
% MixStyle~\cite{zhou2021domain} randomly mixes intermediate feature statistics of different samples across source domains to achieve feature-level enhancement.
\textcolor{black!70!black}{The representation learning methods~\cite{chen2023meta,jo2023poem,mahajan2021domain,akuzawa2020adversarial,luo2024grounding} extract domain-invariant features and learn to separate semantic information for classification.}
% LADG~\cite{9878374} utilizes localized classifiers as domain discriminators in adversarial domain training to achieve fine-grained alignment, while introducing coding-rate loss to mitigate feature space collapse.
% DDG~\cite{Zhang2021TowardsPD} jointly learns a semantic encoder and a variation encoder to disentangle the semantic latent variables and the domain latent variables in data.
% Chen \emph{et al.}~\cite{chen2023meta} learn domain-invariant representations by analyzing and mitigating causal factors that cause domain shift through counterfactual reasoning.
The learning strategy methods~\cite{huang2020self, shi2022gradient, chu2022dna, wang2023sharpness} modify the optimization process to improve generalization.
% RSC~\cite{huang2020self} discards representations related to higher gradients in each round of training and forces the model to use the remaining feature information for prediction.
% DNA~\cite{chu2022dna} theoretically and experimentally links DG with classifier ensemble, verifying that classifier ensemble brings better generalization performance.
% Wang \emph{et al.}~\cite{wang2023sharpness} propose two conditions that can ensure the model to converge to a flat region with improved generalization ability, and propose SAGM that satisfies these two conditions, adapting SAM~\cite{foret2020sharpness} to DG.

\subsection{CLIP-based Domain Generalization}
% Traditional domain generalization methods primarily include data augmentation and representation learning. The data augmentation methods~\cite{wang2021learning,zhou2020learning,nam2021reducing,qiao2020learning} diversify the source domain by perturbing original data and expanding the data distribution within the source domain. 
% The representation learning methods~\cite{chen2023meta,jo2023poem,mahajan2021domain,akuzawa2020adversarial,lv2022causality} extract domain-invariant features and learn to separate semantic information for classification. 
Large-scale pre-trained models~\cite{Devlin2019BERTPO, radford2021learning, rombach2022high} have emerged as powerful tools in modern machine learning, achieving impressive generalization across diverse tasks and domains. By training extensively on vast amounts of data, these models output expressive representations, making them particularly well-suited for tasks that require adaptation to new or unseen scenarios.

Recently, the large-scale pre-trained model, Contrastive Language-Image Pre-Training (CLIP)~\cite{radford2021learning}, has achieved 
impressive performance in various tasks such as image classification~\cite{wang2024hard, tang2024data, cheng2024disentangled, xing2023dual}, image-text retrieval~\cite{tang2024captions,yang2024embracing}, visual question answering~\cite{tschannen2023clippo,merullo2023linearly}, and person re-identification~\cite{he2024exploring}, providing a promising option for domain generalization.
Many methods design prompt tuning schemes for domain generalization with CLIP. 
Zhang \emph{et al.}~\cite{zhang2021amortized} propose a textual prompt learning framework, which constructs batch-wise text prompts based on visual features. 
Bose \emph{et al.}~\cite{bose2024stylip} construct sample-level text prompts using feature statistics and introduce projectors to incorporate intermediate image features into text features. 
Niu \emph{et al.}~\cite{niu2022domain} introduce a source-free manner for CLIP, which contrastively pulls text prompts of the same class across domains closer and separates text prompts of different categories. 
In addition to those methods focusing on text prompts, Li \emph{et al.}~\cite{li2022learning} perform visual prompt tuning (VPT)~\cite{jia2022visual} at each layer of the vision transformer (ViT)~\cite{dosovitskiy2021an} and extend VPT by adding sample-specific tokens mapped from the input context tokens.
% In contrast to these methods focused on prompt learning, our method utilizes the attention refocusing strategy to fine-tune the image encoder of CLIP, enabling CLIP to concentrate on domain-invariant features, enhancing performance for domain generalization.

\subsection{Attention Refocusing in Transformer}
Transformers~\cite{Vaswani2017AttentionIA} are widely used in both natural language processing and computer vision due to their ability to model complex relationships in data.
The attention mechanism plays an important role in Transformers, enabling the model to focus on the most relevant parts of the input and making Transformers effective for a variety of tasks.
Attention refocusing refers to adjusting the attention mechanism in Transformer-based networks based on input data. This process typically involves using additional attention masks, fine-tuning attention weights, or learning a bias for the attention mechanism to emphasize the visual or textual regions of interest. 
This concept has been used in computer vision and neural language processing tasks, such as image classification~\cite{shi2023toast,zhong2022regionclip}, image synthesis~\cite{phung2024grounded}, image segmentation~\cite{xu2023masqclip,xu2023open}, and human-model interaction~\cite{zhang2023tell,sun2024alpha}. 
Phung \emph{et al.}~\cite{phung2024grounded} design a cross-attention loss and a self-attention loss to update the weights of the attention mechanism to adjust the focus on correct image regions. 
Shi \emph{et al.}~\cite{shi2023toast} propose a top-down attention fine-tuning framework to learn bias to adjust the input of value matrices in self-attention.

Although attention refocusing has achieved certain success in various tasks, its capability in domain generalization has not been fully exploited.
Existing studies primarily focus on optimizing attention for known tasks or domains, leaving the question of how to adaptively refocus attention for unseen domains unanswered. This paper introduces attention refocusing in CLIP and proposes a novel framework to adaptively adjust attention in CLIP based on data distributions. Our method not only extends the utility of attention refocusing but also demonstrates its effectiveness in enhancing model generalization across various unknown domains.
% and stabilizes training via a dual-view attention refocusing module and a dual-view parameter ensembling module, respectively. %, since attention refocusing may suffer from the problems of misfitting. 

\section{Our Method}
% \subsection{Formulation}
\begin{figure*}[tb]
	\centering
	% \subfigure[The overall framework of the proposed Dual-view CLIP]{}
	\includegraphics[width=0.8\textwidth]{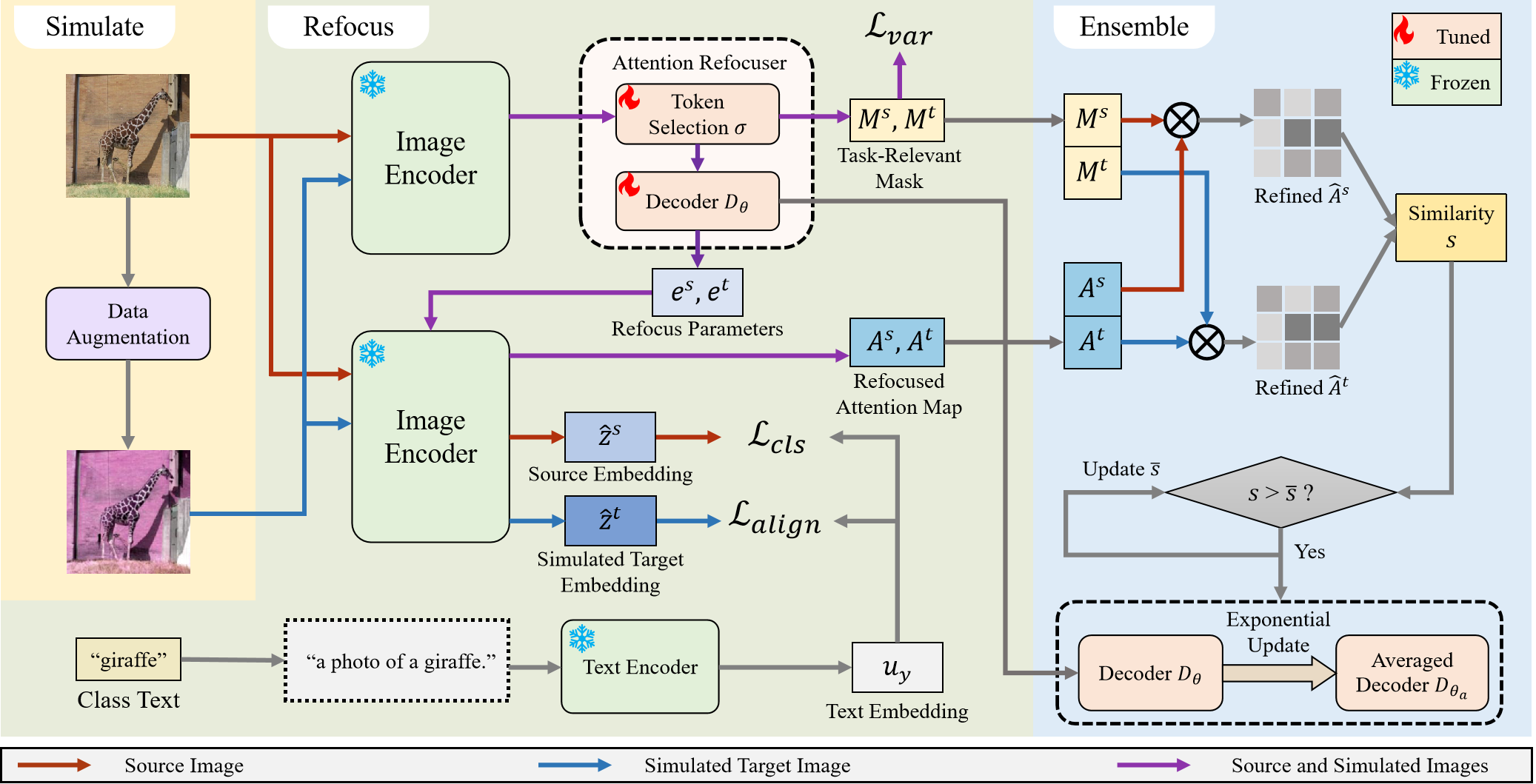}
	% \qquad\quad
	% \subfigure[Dual-view parameter ensembling component]{\includegraphics[width=0.213\textwidth]{img/refined attention.png}}
	\caption{
    Overview of the three stages of the proposed SRE. The \emph{Simulate} stage performs data augmentation to simulate domain shift. The \emph{Refocus} stage introduces an attention-refocuser and learns to align the attention maps of the source and the simulated target images. The \emph{Ensemble} stage selects and ensembles parameters of the attention-refocuser that captures domain-invariant attention maps between the source and the simulated target images.
  %   An overview of the training and inference procedures of the proposed dual-view CLIP that consists of two modules. 
		% The dual-view attention refocusing module comprises a dual-view augmentation and two forward processes. The dual-view parameter ensembling module incorporates the task-relevant mask to refine the regions of interest of CLIP.\jin{what B refer in figure.}
        }
	\label{fig:framework}
\end{figure*}
% There are $K$ source domains available during training, denoted by  $\mathcal{S}_{k}=\{(x_i^{s_k},y_i^{s_k})|_{i=1}^{N_{k}}\}_{k=1}^{K}$
% where $k\in\left\{1,\cdots, K\right\}$, $s_k$ means the $k$-th source domain, $x^{s_k} \in \mathcal{X}^k$ is the image, $y^{s_k} \in \mathcal{Y}^k$ is its label, and there are $N_k$ images in the $k$-th domain totally.
% Each domain $\mathcal{S}_{k}$ is considered to be sampled from a joint probability distribution $P_{\mathcal{XY}}^k$, indicating by $\mathcal{S}_{k} \in P_{\mathcal{XY}}^k$. 
% This paper primarily focuses on the multi-source domain generalization and $P_{\mathcal{XY}}^{k}\neq P_{\mathcal{XY}}^{k'}$ for any $k\neq k^\prime$. 
% The objective of multi-source domain generalization can be expressed as using $K$ source domains to learn a predictive model $f: \mathcal{X}\rightarrow \mathcal{Y}$ that minimizes the prediction error on the target domain $\mathcal{T}=\{x_i^\mathcal{T}|_{i=1}^{N_t}\}$, where $P_{\mathcal{XY}}^{\mathcal{T}}\neq P_{\mathcal{XY}}^{k}$ for all $k\in\left\{1,\cdots, K\right\}$ and the target domain $\mathcal{T}$ is inaccessible during training.
% For clarity, we omit the index $i$ and $k$ for $x_i^{s_k}$ and $y_i^{s_k}$ in the following sections and use $x^s$ to represent source data unless a clear distinction is necessary.

\subsection{Overview}
In this paper, we focus on the problem of multi-source domain generalization, where multiple source domains $\mathcal{S}$ are available during training, while the target domains $\mathcal{T}$ have distributions differing from the source domains and are inaccessible during training.
To tackle this problem, we propose \emph{Simulate, Refocus and Ensemble (SRE)}, a three-stage scheme for multi-source domain generalization, which learns to reduce domain shift by aligning the attention maps in CLIP via attention refocusing.
In the \emph{Simulate} stage, we augment each source image through data transformations to generate simulated target domains, mimicking domain shifts between source and target domains.
In the \emph{Refocus} stage, we introduce an attention-refocuser that simultaneously adjusts the attention of source and simulated target images, mitigating the simulated domain shifts and helping CLIP focus on domain-invariant regions.
In the \emph{Ensemble} stage, we stabilize the training process by selecting and combining robust parameters of the attention-refocuser while removing parameters that are sensitive to the domain shifts, and thus ensure that the focused regions are similar between the source and simulated target images. %which leads to similar focused regions of the source and the simulated target images while removing parameters sensitive to the domain shifts. 
Figure~\ref{fig:framework} illustrates the overview of our method.

\subsection{Simulate via Data Transformation}
%为什么simulate
\textcolor{black}{In multi-source domain generalization, domain shifts between the source and target domains, such as differences in lighting, texture, and color distribution, pose a significant challenge to generalization. Since the target domain data is unavailable during training, our goal is not to precisely simulate the specific domain shifts between the source and target domains, which is inherently infeasible due to the unknown distribution of the target domain. Instead, we aim to create a training scenario that exposes the model to diverse domain shifts, thereby forcing the model to learn to address the domain shifts and enhancing its ability to generalize across unseen domains.}

% In multi-source domain generalization,  domain shifts between the source and target domains, such as differences in lighting, texture, and color distribution, pose a significant challenge to generalization. 
% Since the target data is not available at training time, 
To achieve this, we simulate different degrees of domain shift to diversify the training data as much as possible.
At the same time, to refocus CLIP’s attention on domain-invariant regions, we create simulated target domains $\mathcal{S}^t$ with controlled variability by performing data transformations on source images without changing their spatial location information.
This allows us to model domain shifts during training and ensures that the simulated target images retain the same task-relevant regions as the source images.

The augmentation process is formalized as a sequence of transformations applied to the source image \(x^s\). One transformation is represented as a function \(F_i(x^s, \phi_i)\), where \(\phi_i\) is a parameter controlling the augmentation intensity, uniformly sampled from a predefined range. Given the input source image \(x^s\), we apply three transformations to \(x^s\) and the corresponding simulated target image \(x^t\)  is given by
\begin{equation}
x^t = F_3(F_2(F_1(x^s, \phi_1), \phi_2), \phi_3),
\end{equation}  
where \(F_1\), \(F_2\), and \(F_3\) represent specific augmentation operations. In our implementation, the three transformations are \textbf{ColorJitter}, \textbf{GaussianBlur}, and \textbf{GrayScale}.
\textbf{ColorJitter} adjusts brightness, contrast, saturation, and hue to simulate lighting changes; \textbf{GaussianBlur} introduces slight blurring to mimic variations in focus; \textbf{GrayScale} reduces color information to simulate grayscale conditions. By using the three transformations, the simulated target domain \(\mathcal{S}^t\) is created 
% as \(\mathcal{S}' \leftarrow \text{Augment}(\mathcal{S})\), 
to mimic the distribution discrepancy between source and target domains.
%and the distribution discrepancy between source and target doamins is mimiced.

% To simulate the domain shifts between source domains and target domains, we apply data transformation to each source image $x$ to generate its augmented view $x'$, forming a simulated domain $\mathcal{S}'\leftarrow \mathrm{Augment}(\mathcal{S})$. The distribution discrepancy between source images and augmented views serves as the simulated domain shift. 
% Since our scheme aims to focus on domain-invariant regions, we augment each original image $x$ without changing its spatial position, ensuring that the task-relevant regions between the original image and its augmented view are consistent.
% The augmentation specifically includes ColorJitter, Random GaussianBlur, and Random GrayScale.

\subsection{Refocus via Attention Map Alignment}
\textcolor{black}{A recent study~\cite{yao2024theoretical} provides a theoretical analysis that fine-tuning the query or value matrices within the attention mechanism can improve generalization bounds and enhance memory efficiency. The goal of domain generalization is to develop models that can perform effectively on unseen target domains by leveraging knowledge from multiple source domains. Attention mechanisms contribute to this goal by emphasizing domain-invariant features while minimizing reliance on domain-specific attributes~\cite{meng2022attention}.}

\begin{figure}[tb]
	\centering
	% \subfigure[The overall framework of the proposed Dual-view CLIP]{}
	\includegraphics[width=\linewidth]{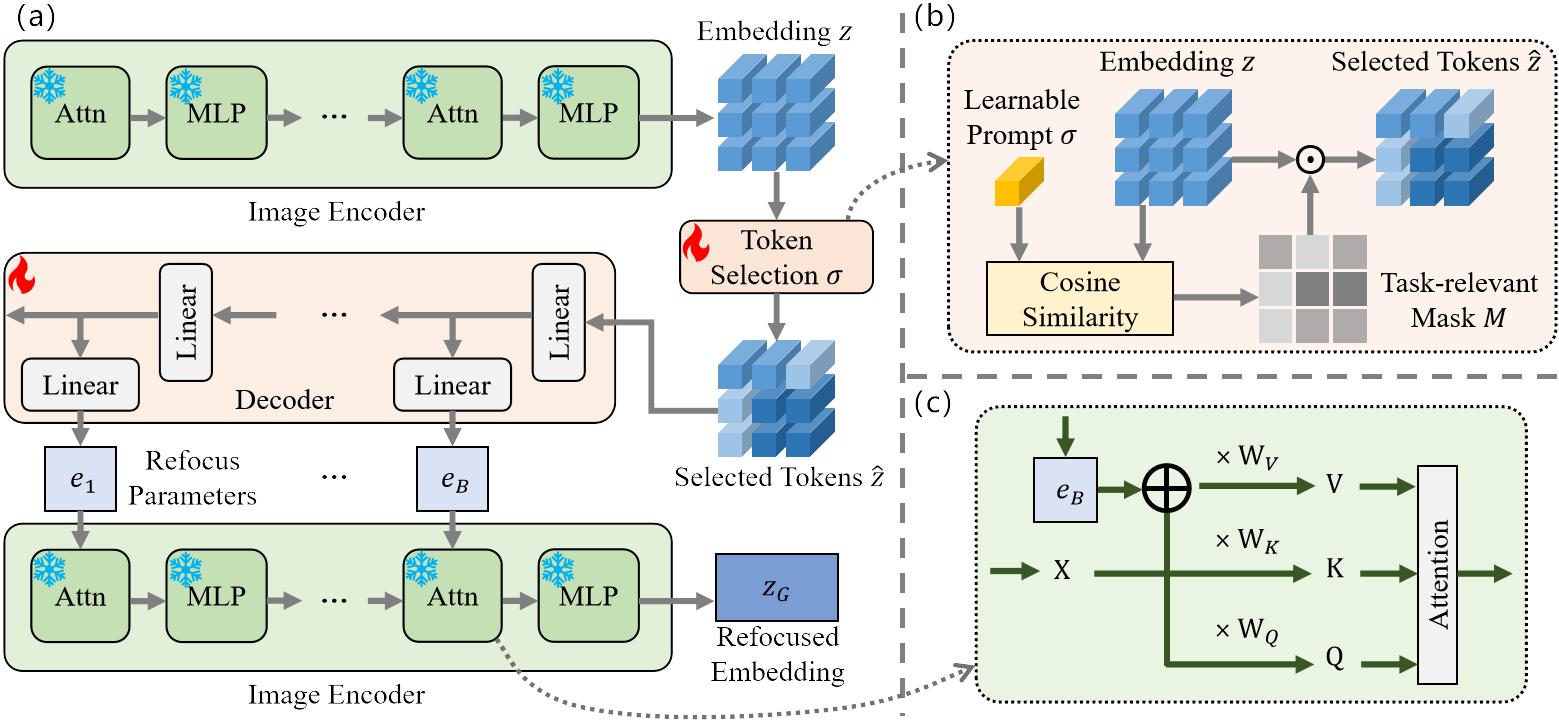}
	% \qquad\quad
	% \subfigure[Dual-view parameter ensembling component]{\includegraphics[width=0.213\textwidth]{img/refined attention.png}}
	\caption{The architecture of the attention-refocuser in the \emph{Refocus} stage. (a) The attention-refocuser consists of a token selection module and a decoder. (b) The token selection module uses a learnable prompt to select task-relevant tokens. (c) The modified attention mechanism adds the refocus parameters to the input of the value matrix calculation.}
	\label{fig:decoder}
\end{figure}

Given the source domains and the simulated target domains, we introduce an attention-refocuser to bridge the domain shifts between them. Specifically, the attention-refocuser learns to adaptively refine attention maps of source images and simulated target images based on their feature distributions, through which attention maps of simulated target images are aligned with source images. We display the detailed architecture of the attention-refocuser in Figure~\ref{fig:decoder}.

Denote $f_v$ as the image encoder of CLIP and $E$ as the original attention parameter in $f_v$, the attention-refocuser $R_g$ learns to modify the attention parameter $E$ in $f_v$ to reduce the domain shifts between source domains and simulated target domains. $g= \{ \sigma, \theta\}$ is the parameter of the attention-refocuser $R_g$, consisting of a learnable prompt $\sigma$ for a token selection module and an attention decoder $D_\theta$ parameterized by $\theta$.
The attention parameter $E$ in $f_v$ for a source image $x^s$ is refined based on its visual embedding $f_v(x^s,E)$, denoted by $\hat{E}^s= \left\{E, e^s\right\}$, where $e^s=R_g(f_v(x^s,E))$ is the newly added parameter for the attention mechanism.
% The modified attention parameters in $f_v$ is denoted by $E'=\left\{E,g\right\}$.
% the modified attention mechanism in $f_v$ integrates the refocus embeddings $E=G(f_v(x))$ so that 
Similarly,  the refined attention parameter for the simulated target image $x^t$ is given by $\hat{E}^t= \left\{E, e^t\right\} = \left\{E, R_g(f_v(x^t, E)) \right\}$. 
In this case, CLIP extracts the domain-invariant embeddings $\hat{z}^s$ and $\hat{z}^t$ of the source image $x^s$ and the simulated target image $x^t$, respectively, formulated by
\begin{equation}
	\label{equ:forward}
        \begin{aligned}
	\hat{z}^s = f_v\Big(x^s, \hat{E}^s \Big), \ 
	\hat{z}^t = f_v\Big(x^t, \hat{E}^t \Big).
    \end{aligned}
\end{equation}
Specifically, both the source image and its simulated target image are passed through two forward processes of $f_v$.
% to update the attention via the attention-refocuser $R_g$.
In the first forward process, visual embeddings $z^s=f_v(x^s, E)$ and $z^t=f_v(x^t, E)$ are extracted from $x^s$ and $ x^t$  and then used to generate refocus parameters $e^s$ and $e^t$ for each self-attention layer. In the second forward process, 
$e^s$ and $e^t$ are used in turn to adjust the self-attention mechanism in $f_v$, yielding the refocused visual embeddings $\hat{z}^s$ and $\hat{z}^t$ for the source image and the simulated target image, respectively.
% The dual-view attention refocusing module is shown in Figure~\ref{fig:framework}. 

% Concretely, the weak augmentation process involves randomly cropping the original image and resizing it to a predefined size, along with a random horizontal flip, resulting in the weakly augmented view $x'$.
% For the strong augmentation, $x'$ undergoes further augmentation through a mixture of additional transformations, including color jitter, Gaussian blur, and grayscale adjustments, ultimately generating the strongly augmented view $x''$.

% When it comes to the first forward process, the image encoder $f_v$ receives the two views $x'$ and $x''$ as input and produces $z', z''$ of the two views which consists of $L$ tokens $z' = \left\{z'_l\right\}_{l=1}^L$, $z'' = \left\{z''_l\right\}_{l=1}^L$ respectively, where $z'_l, z''_l \in \mathbb{R}^d$, $z', z'' \in \mathbb{R}^{L\times d}$, and $d$ represents the embedding dimension of the transformer. 
During the first forward process, the image encoder $f_v$ takes the source image $x^s$ and the simulated target image $x^t$ as input and produces corresponding visual embeddings $z^s$ and $z^t$, consisting of $L$ tokens each, represented as $z^s = \left\{z^s_l\right\}_{l=1}^L$ and $z^t = \left\{z^t_l\right\}_{l=1}^L$, where $z^s_l, z^t_l \in \mathbb{R}^d$, $z^s, z^t \in \mathbb{R}^{L\times d}$, with $d$ representing the embedding dimension of the transformer.
Then, we feed $z^s, z^t$ into the attention-refocuser $R_g$ which consists of a token selection module and a decoder. The token selection module extracts task-relevant tokens from $z^s, z^t$ by
calculating two task-relevant masks $M^s, M^t\in \mathbb{R}^L$ using a learnable task-relevant prompt $\sigma \in \mathbb{R}^d$:
% using a learnable task-relevant prompt $\sigma \in \mathbb{R}^d$ and 
\begin{equation}
	\label{sim_mat}
	M^s_l=\left<z^s_l , \sigma\right>, \ \  M^t_l=\left<z^t_l , \sigma\right>, 
\end{equation}
where $M^s_l$ and $M^t_l$ denote the $l$-th element in $M^s$ and $M^t$, respectively, measuring the task relevance of the $l$-th token in $z^s$ and $z^t$. $\left<\cdot,\cdot \right>$ is the cosine similarity.
Selected tokens are denoted by $\tilde{z}^s = (M^s \odot z^s)$ and $\tilde{z}^t = (M^t \odot z^t)$, where $\odot$ is the Hadamard product. 
To increase the disparity of the task-relevant mask at different token positions, we further compute a variance loss on $M^s$ and $M^t$:
\begin{equation}
	\label{loss_var}
	\mathcal{L}_{var}= \mathbb{E}\left[(M^s - \mathbb{E}[M^s])^2\right] + \mathbb{E}\left[(M^t - \mathbb{E}[M^t])^2\right].
\end{equation}
% where $Var$ denotes the variance calculation of $M^s$ or $M^t$.

% The channel selection is performed over the results of token selection using a transform matrix $\mathbf{P} \in \mathbb{R}^{d\times d}$. The final output of the entire feature selection process is computed as $\hat{z}' = (M' \odot z')\cdot \mathbf{P}$ and $\hat{z}'' = (M'' \odot z'')\cdot \mathbf{P}$.

The decoder $D_\theta$ comprises $B$ layers, matching the number of layers in $f_v$, to convert the selected tokens into refocus parameters for attention. 
% Each decoder layer consists of a linear transformation. 
Given $\tilde{z}^s$ as the input for the first layer of $D_\theta$, 
% The decoder $g$ in the feedback path consists of $B$ layers, matching the number of layers in CLIP's vision encoder. Each decoder layer consists of a linear transformation applied to the previous output, where $\hat{z}_l$ serves as the input to the first decoder layer. 
the output of the last $b$-th layer in the decoder $D_\theta$ is $e^s_b \in \mathbb{R}^{L\times d}$, which is used as the refocus parameter to adjust the attention mechanism in the $b$-th layer in $f_v$.
All outputs of the decoder are represented as $e^s = \left\{e^s_b\right\}_{b=1}^B$. The refocus parameters $e^t$ for $\tilde{z}^t$ can be obtained with the same procedure.

When it comes to the second forward process, the self-attention mechanism within each layer in the image encoder $f_v$ integrates the corresponding refocus parameters into the computation of the value matrix.
Conventionally, given the input $x^s$, the value matrix $\mathbf{V}_b$ in the $b$-th layer is derived as $\mathbf{V}_b = \mathbf{W}_{V_b}\mathbf{X}_b$, where $\mathbf{W}_{V_b}$ denotes the weight matrix corresponding to the value operation and $\mathbf{X}_b$ denotes the input visual embedding corresponding to $x^s$ in the $b$-th layer.
%In contrast, during the second forward process in our method, 
The refocus parameters $e^s_b$ serve as a bias, modifying the calculation to $\hat{\mathbf{V}}_b = \mathbf{W}_{V_b}(\mathbf{X}_b + e^s_b)$, where $\hat{\mathbf{V}}_b$ represents the value matrix adjusted by the refocus parameters $e^s_b$. So we obtain the refined attention parameters $\hat{E}^s = \left\{E, e^s\right\}$ and the refocused visual embeddings $\hat{z}^s=f_v(x^s, \hat{E}^s)$. Similarly, we derive $\hat{z}^t=f_v(x^t,\hat{E}^t)$ for the simulated target images.

We use the text encoder $f_t$ of CLIP to obtain text embeddings $u_c = f_t(t_c)$
% \begin{equation}
% \label{text_feat}
%     u_c = f_t(t_c)
% \end{equation}
for each category $c$, where $t_c$ is the text prompt, constructed as $t_c = \mathrm{``a\ photo\ of\ a\ } [CLASS(c)]\mathrm{"}$, 
% \begin{equation}
% 	\label{text_prompt}
% 	t_c = \mathrm{``a\ photo\ of\ a\ } [CLASS(c)]\mathrm{"}, 
% \end{equation}
following the same manner in zero-shot CLIP~\cite{radford2021learning}.
With the incorporation of attention refocusing, the prediction probabilities of the source image and the simulated target image for the label $y$ are given by
% The second forward pass considers both the original image and the top-down signals as input and outputs the final features $f_v(x, \mathbf{X}^\mathrm{td})$. With top-down attention, the prediction probability becomes:
\begin{equation}
    \label{equ:pred}
    \begin{split}
        p(y|x^s) &= \frac{\exp(\left<u_y, \hat{z}^s\right>/\tau)}{\sum\limits_{c\in \mathcal{Y}}\exp(\left<u_c, \hat{z}^s\right>/\tau)}, \\
        p(y|x^t) &= \frac{\exp(\left<u_y, \hat{z}^t\right>/\tau)}{\sum\limits_{c\in \mathcal{Y}}\exp(\left<u_c, \hat{z}^t\right>/\tau)},
    \end{split}
\end{equation}
where $\left<\cdot,\cdot \right>$ denotes cosine similarity and $\tau$ is a temperature coefficient.
% \begin{equation}
	% \label{text_prompt}
	%     t_y = \text{"a photo of a } [CLASS(y)]\text{."}
	% \end{equation}
% following the same manner in zero-shot CLIP~\cite{radford2021learning}.
We then employ the cross-entropy loss function on the source domains $\mathcal{S}$ to optimize the attention-refocuser, formulated by 
\begin{equation}
	\label{loss_ref}
	\mathcal{L}_{cls} =  \mathbb{E}_{(x^s,y)\sim\mathcal{S}} \big[ -\log p(y|x^s) \big],
\end{equation}
while forcing the attention-refocuser to simultaneously correctly refocus the attention map of the simulated target image, thereby aligning the source domain $\mathcal{S}$ and simulated target domain $\mathcal{S}^t$:
\begin{equation}
	\label{loss_ce}
	\mathcal{L}_{align} =  \mathbb{E}_{(x^t,y)\sim\mathcal{S}^t} \big[ -\log p(y|x^t) \big].
\end{equation}
Consequently, the parameters $R=\{\sigma, \theta\}$ of the attention-refocuser $R_g$ are jointly optimized by the total loss:
\begin{equation}
\label{equ:loss_all}
    \mathcal{L}=  \mathcal{L}_{cls}+ \mathcal{L}_{align}+ \lambda \mathcal{L}_{var},
\end{equation}
where  $\lambda$ is a scaling hyper-parameter.
% At each updating step, the total loss is averaged by the number of samples in one batch $\mathcal{B}$, denoted as $\overline{\mathcal{L}} = \frac{1}{|\mathcal{B}|}\sum_{i=1}^{|\mathcal{B}|}\mathcal{L}_i$, where $\mathcal{L}_i$ is the total loss for the $i$-th sample in the current batch. 
% Finally the total loss is used to optimize the parameters $R=\{\sigma, \theta\}$ of the attention refocuser.

% We optimize the parameters $R=\{\sigma, \theta\}$ of the attention refocuser by
% \begin{equation}
% 	\label{equ:update}
% 	\{\sigma, \theta\}\leftarrow \{\sigma, \theta\}-\eta\nabla\overline{\mathcal{L}},
% \end{equation}
% where $\nabla\overline{\mathcal{L}}$ is the gradient for $\{\sigma, \theta\}$ and $\eta$ is the learning rate.

\subsection{Ensemble via  Parameter Selection}
Although the \emph{Refocus} stage is capable of aligning source domains and simulated target domains, it may also suffer from the problem of training instability. Instead of explicitly aligning the attention maps of source images and simulated target images, which may lead to over-fitting, we propose to ensemble parameters of the attention-refocuser to generate domain-invariant attention maps. %between the source image and the simulated target image, preventing training instability. 

% Based on the Simulate and Refocus stages, we propose an Ensemble stage that selects parameters of the attention refocuser, which performs consistent focus between the source image and the augmented view, preventing training instability. 
Considering that CLIP relies solely on class embedding in the last layer for classification, the attention map in the last layer contains much semantic information. Therefore, we use it to represent the focus of CLIP, denoted as $A = \mathrm{softmax}(\frac{\mathbf{Q}_{B}\mathbf{K}_{B}^T}{\sqrt{d_k}})$, where $\frac{1}{\sqrt{d_k}}$ serves as a scaling factor. 
% only the attention map over the class embeddings, which denotes as $A$, is crucial to be considered. 
% We regard the focus of CLIP as the attention map in the last self-attention layer, denoted as $Attn = \mathrm{softmax}(\frac{\mathbf{Q}_{B}\mathbf{K}_{B}^T}{\sqrt{d_k}})$, where $\frac{1}{\sqrt{d_k}}$ serves as a scaling factor. 
We extract the attention maps $A^s$ and $A^t$ of to effectively represent the focus of CLIP on the source image $x^s$  and the simulated target image $x^t$, respectively.
Then, we use the task-relevant masks $M^s$ and $M^t$ calculated in Eq.~\eqref{sim_mat}  to refine the attention maps $A^s$ and $A^t$, respectively, better representing the regions of interest in CLIP on the source image and the simulated target image.
% To better represent the regions of interest in our model, we introduce the task-relevance mask $M', M''$ calculated in Eq.~\ref{sim_mat} which display a rough focus of the model to refine the attention map $A'$, better representing the task-relevant regions of an image.
In this case, the refined attention maps are computed as 
\begin{equation}
	\label{equ:refine}
	\hat{A}^s = A^s \odot M^s,
\end{equation}
while the refined regions of interest $\hat{A}^t$ for the simulated target image $x^t$ are calculated in the same way.

To stabilize the training procedure, we select the parameters $\theta$ of the decoder $D_{\theta}$ in the attention-refocuser that leads to high consistency between $\hat{A}^s$ and  $\hat{A}^t$. The consistency is measured by  the cosine similarity between the two attention maps, formalized as
\begin{equation}
	\label{equ:similarity}
	s = \left<\hat{A}^s,\hat{A}^t\right>.
\end{equation}

We maintain an exponential updated score $\overline{s}$ of the similarity as a threshold to select the parameters $\theta$. At the $t$-th step, when the average similarity $s_t$ of all cosine similarities $s$ in the current batch is larger than the threshold $\overline{s}$, we ensemble the current parameter $\theta_t$ of the decoder and simultaneously update the threshold using
% \begin{equation}
% 	\label{equ:ratio}
% 	\theta_{a} \leftarrow \omega \theta_{a} + (1-\omega) \theta_t,
% 	\ \ \hat{s} \leftarrow \omega \hat{s} + (1-\omega) s_t,
% \end{equation}
\begin{equation}
\label{equ:ratio1}
	\theta_{a}\leftarrow \left\{
	\begin{array}{lr}
	\omega \theta_{a} + (1-\omega) \theta_t,& s_t>\overline{s}\\
	\theta_{a},& s_t\leq\overline{s}
	\end{array}\right.,
	% \theta_{a} \leftarrow \omega \theta_{a} + (1-\omega) \theta_t,
	%\ \ \hat{s} \leftarrow \omega \hat{s} + (1-\omega) s_t,
\end{equation}
\begin{equation}
\label{equ:ratio2}
	\overline{s} \leftarrow \left\{
	\begin{array}{lr}
	\omega \overline{s} + (1-\omega) s_t, & s_t>\overline{s}\\
	\overline{s},& s_t\leq\overline{s}
	\end{array}\right.,
	% \theta_{a} \leftarrow \omega \theta_{a} + (1-\omega) \theta_t,
	%\ \ \hat{s} \leftarrow \omega \hat{s} + (1-\omega) s_t,
\end{equation}
% \sigma^{a} &\leftarrow \omega \sigma^{a} + (1-\omega) \sigma \label{ema:1},\\
% \mathbf{P}^{a} &\leftarrow \omega \mathbf{P}^{a} + (1-\omega) \mathbf{P} \label{ema:2},\\
% \end{align}
where $\overline{s}$ is initialized to zero, and $\omega$ is the update ratio for the ensembled parameter $\theta_a$ and the threshold $\overline{s}$. The ensembled parameter $\theta_a$ is initialized to all zeros. The ensembled decoder $D_{\theta_{a}}$ will be used in the inference stage. 

\subsection{Inference}
During testing, each target sample $x^{t}$ is fed into the image encoder $f_v$ of CLIP to obtain its original visual embedding $z^{\mathcal{T}}$. 
Then the attention-refocuser adaptively adjusts attention through the task-relevant prompt $\sigma$ and the ensembled decoder $D_{\theta_a}$ to produce the refocus parameters $e^{\mathcal{T}}$. 
The second forward process inputs the original test sample $x^{\mathcal{T}}$ and the refined parameters $E^{\mathcal{T}}=\left\{E,e^{\mathcal{T}}\right\}$, and outputs the refocused visual embedding $\hat{z}^{\mathcal{T}}$. Finally, $\hat{z}^{\mathcal{T}}$ is compared with text embeddings of all categories, and the category with the highest similarity is selected as the prediction result:
% In addition, the predicted label is obtained by selecting the category with the highest prediction probability:
\begin{equation}
\hat{y} = \arg\max_{c\in \mathcal{Y}} p(c|x^{\mathcal{T}}),
\end{equation}
where $p(c|x^{\mathcal{T}})$ is calculated according to Eq.~\eqref{equ:pred}.

% \begin{figure*}[tb]
% 	\centering
% 	% \subfigure[The overall framework of the proposed Dual-view CLIP]{}
% 	\includegraphics[width=\textwidth]{img/inference.png}
% 	% \qquad\quad
% 	% \subfigure[Dual-view parameter ensembling component]{\includegraphics[width=0.213\textwidth]{img/refined attention.png}}
% 	\caption{The inference procedure of the proposed SRE for a given input image, where the ensembled parameters are used for refocusing the attention in CLIP.}
% 	\label{fig:inference}
% \end{figure*}

\section{Experiments}
\subsection{Datasets}
\label{sec:ds}
We conduct experiments on five commonly used datasets: VLCS~\cite{fang2013unbiased}, PACS~\cite{li2017deeper}, OfficeHome~\cite{venkateswara2017deep}, TerraIncognita~\cite{beery2018recognition} and DomainNet~\cite{peng2019moment}.
{VLCS} consists of 10,729 images sourced from five distinct classes originating from four separate datasets: PASCAL VOC 2007~\cite{everingham2010pascal}, LabelMe~\cite{russell2008labelme}, Caltech~\cite{fei2004learning}, and Sun~\cite{xiao2010sun}. 
% Each of these datasets is treated as an individual domain in the domain generalization context.
{PACS} contains 9,991 images classified into seven categories, exhibiting significant distribution shifts across four domains: art painting, cartoon, photo, and sketch.
{OfficeHome} contains 15,500 images across 65 categories in both office and home environments. These categories in OfficeHome are distributed among four domains: Art, Clipart, Product, and Real-World, characterized by distinct viewpoints and image styles.
{TerraIncognita} (Terra) contains 24,788 animal images captured in the wild across different locations. It encompasses a total of ten animal classes and each location is treated as a distinct domain, denoted as Location100, Location38, Location43, and Location46.
{DomainNet} contains 586,575 images across 345 classes, categorized into six types of domains: clipart, infograph, painting, quickdraw, real, and sketch.

\subsection{Implementation Details}
\label{sec:implement}
% Our code is implemented based on DomainBed~\footnote{\url{https://github.com/facebookresearch/DomainBed}}, a widely used benchmark for domain generalization. Consistent with prior studies, 
We use ViT-B/16~\cite{dosovitskiy2021an}, ViT-B/32~\cite{dosovitskiy2021an} and ViT-L/14~\cite{dosovitskiy2021an} as the backbone for CLIP. We train the parameters $g$ of the attention-refocuser via 2000 iterations for PACS, VLCS, and OfficeHome, while 5000 iterations for Terra and DomainNet. The AdamW optimizer is adopted. The learning rate is set to 0.0004 for the decoder parameter $\theta$ and 0.001 for the task-relevant prompt $\sigma$, and the weight decay is set to 0.005 across all datasets. $\theta$ is randomly initialized while $\sigma$ is initialized to constant 1 for all positions. The hyperparameters $\omega$ and $\lambda$ are set to 0.98 and 0.1, respectively, across all datasets. We opt for a mini-batch size of 16 and perform four gradient accumulation steps during training to achieve a larger batch size.
% To perform dual-view augmentation, we specifically apply RandomResizeCrop and RandomHorizontalFlip to generate weakly augmented images and compose ColorJitter, GaussianBlur, and RandomGrayScale for strong augmentation. 
%我目前没有调研到对这些增强操作进行引用的文献，可能因为这些是基础的增强，不是说像RandAugmentation那类专门设计数据增强的方法。
%For the Terra dataset, we incorporate a CosineScheduler to gradually decrease the learning rate to 0.01 of the initial value.
It's noteworthy that we adhere to the training-domain validation strategy~\cite{gulrajani2020search} that utilizes subsets from each training domain for validation to select the best-performing model. We split the data of each training domain into 80\% for training and 20\% for validation, subsequently combining the validation sets of all domains for model selection. All experiments are conducted on a single NVIDIA RTX 4090 GPU with 24GB of graphics memory.
% Additionally, we gradually decrease the learning rate to 1e-6 using a CosineScheduler. Importantly, we follow the training-domain validation set in DomaindBed which indicates using the subsets of each training domain for validation to select a model. Following the default setting in DomainBed, we split the data of each training domain into 0.8 and 0.2 then pool them together for model selection.
% \subsection{Experimental Results}

\subsection{Comparison  with CLIP-based methods}

\begin{table}[tb]
\caption{Comparison results between SRE and other CLIP-based methods on PACS, VLCS, OfficeHome, Terra, and DomainNet for multi-source domain generalization using different backbones. The results of these compared methods are copied by default from their original papers. ``*" denotes our reproduced results, ``$\dagger$" denotes results copied from SPG~\cite{bai2025soft}.}
\label{tab:all_result} 
\centering
	\scalebox{0.9}{\begin{tabular}{l|ccccc|c}
			\toprule
			{\bf Method} & {\bf PACS} & {\bf VLCS}&{\bf Off.H.}&{\bf Terra}&{\bf Dom.N.}&{\bf Avg.}\\
			\hline
            \multicolumn{7}{c}{ViT-32/B}\\
            \hline
			ZS-CLIP~\cite{radford2021learning} & \underline{94.9} & 80.9 & 78.9 & 22.6 & 53.9 & 66.2\\
			VPT*~\cite{jia2022visual}   & \textbf{95.1} & \underline{81.9} & 80.3 & 40.3 & 50.8 & 69.7\\
			MaPLe*~\cite{khattak2023maple}   & 94.6 & 81.6 & \underline{81.3} & \underline{41.5} & \underline{55.8} & \underline{70.9}\\
			\textbf{SRE (Ours)}  & \textbf{95.1} & \textbf{82.8} & \textbf{82.5} & \textbf{49.0} &\textbf{57.3} & \textbf{73.3}\\
			\hline
            \multicolumn{7}{c}{ViT-16/B}\\
            \hline
			ERM~\cite{vapnik1998statistical}  & 92.9 & 81.4 & 78.9 & 53.6 & 56.1 & 72.6\\
			DANN~\cite{ganin2016domain} & 92.2 & 80.1 & 78.0 & 47.9 & 57.5 & 71.1\\
			CORAL~\cite{sun2016deep} & 92.6 & 79.6 & 78.5 & 51.7 & 56.4 & 71.8\\
			MIRO~\cite{cha2022domain} & 95.2 & 81.1 & 82.5 & 52.9 & 56.6 & 73.7\\
			ZS-CLIP~\cite{radford2021learning} & 96.1 & 82.5 & 82.1 & 33.9 & 57.5 & 70.4\\
			DPL\textsuperscript{$\dagger$}~\cite{zhang2021amortized}  & 96.4 & 80.9 & 83.0 & 46.6  & 59.5 & 73.6\\
			VPT\textsuperscript{$\dagger$}~\cite{jia2022visual}   & 96.9 & 82.0 & 83.2 & 46.7 & 58.5 & 73.6\\
			CSVPT~\cite{li2022learning} & 96.6 & 82.7 & 85.5 & \underline{57.7} & 59.8 & 76.5\\
   	        MaPLe\textsuperscript{$\dagger$}~\cite{khattak2023maple}   & 96.5 & 82.2 & 83.4 & 50.2 & 59.5 & 74.4\\
            SPG~\cite{bai2025soft} & \underline{97.0} & 82.4 & 83.6 & 50.2 & 60.1 & 74.7\\
            % CLIP-Adapter~\cite{gao2024clip} & 96.4 & \underline{84.3} & 82.2 & {\bf --} & 59.9 & {\bf --}\\
            % Gallop*~\cite{lafon2024gallop} & 96.6 & 83.8 & \underline{86.0} & 51.7 & \underline{61.8} & 76.0\\
            % CLIPCEIL~\cite{yu2024clipceil} & \textbf{97.6} & \textbf{88.4} & 85.4 & 53.0 & \textbf{62.0} & \underline{77.3}\\
            \textcolor{black}{CLIP-Adapter~\cite{gao2024clip}} & \textcolor{black}{96.4} & \textcolor{black}{\underline{84.3}} & \textcolor{black}{82.2} & \textcolor{black}{{\bf --}} & \textcolor{black}{59.9} & \textcolor{black}{{\bf --}}\\
            \textcolor{black}{Gallop*~\cite{lafon2024gallop}} & \textcolor{black}{96.6} & \textcolor{black}{83.8} & \textcolor{black}{\underline{86.0}} & \textcolor{black}{51.7} & \textcolor{black}{\underline{61.8}} & \textcolor{black}{76.0}\\
            \textcolor{black}{CLIPCEIL~\cite{yu2024clipceil}} & \textcolor{black}{\textbf{97.6}} & \textcolor{black}{\textbf{88.4}} & \textcolor{black}{85.4} & \textcolor{black}{53.0} & \textcolor{black}{\textbf{62.0}} & \textcolor{black}{\underline{77.3}}\\
			\textbf{SRE (Ours)}  & \underline{97.0} & 83.7 & \textbf{86.1} & \textbf{60.8} & \underline{61.8} & \textbf{77.9}\\
            \hline
            \multicolumn{7}{c}{ViT-14/L}\\
            \hline
			ZS-CLIP~\cite{radford2021learning}  & 98.3 & 81.9 & 86.6 & 43.7 & 63.0 & 74.7\\
    		VPT*~\cite{jia2022visual} & 98.0 & \underline{83.2} & \underline{89.4} & 60.4 & 63.9 & 79.0\\	
            CSVPT~\cite{li2022learning} & \underline{98.5} & 83.0 & \textbf{90.9} & \textbf{65.3} & \underline{65.3} & \underline{80.6}\\
            MaPLe*~\cite{khattak2023maple} & 98.3 & 81.9 & 89.0 & 57.0 & 63.5 & 77.9\\
			\textbf{SRE (Ours)} & \textbf{98.7} & \textbf{84.3} & \textbf{90.9} & \underline{64.8} & \textbf{67.0} & \textbf{81.1}\\
			\bottomrule
	\end{tabular}}
\end{table}

\textcolor{black}{We conduct a comparative analysis of our method (SRE) with CLIP-based methods, including the baseline zero-shot CLIP (ZS-CLIP)~\cite{radford2021learning}, parameter-efficient tuning methods (DPL~\cite{zhang2021amortized}, VPT~\cite{jia2022visual}, CSVPT~\cite{li2022learning}, MaPLe~\cite{khattak2023maple}, SPG~\cite{bai2025soft}, CLIP-Adapter~\cite{gao2024clip}, Gallop~\cite{lafon2024gallop}, CLIPCEIL~\cite{yu2024clipceil}), representation learning methods (DANN~\cite{ganin2016domain}, CORAL~\cite{sun2016deep}, MIRO~\cite{cha2022domain}) with their baseline ERM~\cite{vapnik1998statistical}.
% \zhi{why only introduce CSVPT and ZS-CLIP here?}
% CSVPT~\cite{li2022learning} stands out as the state-of-the-art parameter-efficient fine-tuning method for CLIP. 
% Zero-shot CLIP (ZS-CLIP)~\cite{radford2021learning} is a baseline for CLIP-based methods, which applies zero-shot classification to CLIP and constructs a text prompt as "a photo of a [CLASS]" for each category. 
The comparison results on the five datasets are shown in Table~\ref{tab:all_result}. We have the following observations. 
% Our method achieves the best performance with almost all backbones across all datasets,
Our method achieves the overall best performance on almost all backbones,
which validates the effectiveness of the proposed SRE scheme for domain generalization. While CLIPCEIL shows strong results on VLCS, our method's significant lead on the challenging TerraIncognita benchmark (+7.8\%) highlights its enhanced robustness against substantial domain shifts.}
Table~\ref{tab:detailed_pacs}-\ref{tab:detailed_dom} show the detailed comparison results of all domains on the PACS, VLCS, OfficeHome, Terra and DomainNet datasets, respectively, using the ViT-16/B backbone.

\begin{table}[htbp]
    \caption{Comparison results between SRE and other CLIP-based methods on PACS under the leave-one-domain-out setting. ``*" denotes our reproduced results, ``$\dagger$" denotes results copied from SPG~\cite{bai2025soft}
  }
  \label{tab:detailed_pacs}
  \centering
    \scalebox{0.9}{\begin{tabular}{l|p{1.cm}<{\centering}|p{1.cm}<{\centering}|p{1.cm}<{\centering}|p{1.cm}<{\centering}|p{0.6cm}<{\centering}}
    \toprule
    {\bf Method} & {\bf Art} & {\bf Cartoon}&{\bf Photo}&{\bf Sketch}& {\bf Avg.}\\ \midrule
    % \midrule
    ZS-CLIP~\cite{radford2021learning} & 97.1 & \textbf{99.1} & \textbf{99.9}  & 88.1 & 96.1\\
    DPL\textsuperscript{$\dagger$}~\cite{zhang2021amortized}  & 97.8 & 98.5 & \textbf{99.9} & 89.5 & 96.4\\
    VPT\textsuperscript{$\dagger$}~\cite{jia2022visual}  & 97.9 & 98.9 & \textbf{99.9} & 91.0 & 96.9\\
    MaPLe\textsuperscript{$\dagger$}~\cite{khattak2023maple} & 97.9 & 98.7 & 99.7 & 89.8 & 96.5 \\
    % PromptSRC*~\cite{khattak2023self} & 97.9 & 99.1 & \textbf{99.9} & 89.9 & 96.7 \\
    SPG~\cite{bai2025soft}  & 97.7 & 99.0 & \textbf{99.9} & \textbf{91.3} & \textbf{97.0}\\
    % Gallop*~\cite{lafon2024gallop}  & \textbf{98.3} & 98.9 & \textbf{99.9} & 89.3 & 96.6\\
    \textcolor{black}{Gallop*~\cite{lafon2024gallop}} & \textcolor{black}{\textbf{98.3}} & \textcolor{black}{98.9} & \textcolor{black}{\textbf{99.9}} & \textcolor{black}{89.3} & \textcolor{black}{96.6}\\
    \textbf{SRE (Ours)}  & \textbf{98.3}  & 99.0  & 99.8 & 91.1 & \textbf{97.0}\\
    \bottomrule
\end{tabular}}
\end{table}
\vspace{-5mm}
\begin{table}[htbp]
\caption{Comparison results between SRE and other CLIP-based methods on VLCS  under the leave-one-domain-out setting. ``*" denotes our reproduced results, ``$\dagger$" denotes results copied from SPG~\cite{bai2025soft}
  }
  \label{tab:detailed_vlcs}
  \centering
\scalebox{0.9}{\begin{tabular}{l|p{1.0cm}<{\centering}|p{1.0cm}<{\centering}|p{1.0cm}<{\centering}|p{1.0cm}<{\centering}|p{0.6cm}<{\centering}}
    \toprule
    {\bf Method} & {\bf Caltech} & {\bf LabelMe}&{\bf SUN}&{\bf VOC}& {\bf Avg.}\\ \midrule
    % \midrule
    ZS-CLIP~\cite{radford2021learning} & 99.8 & \textbf{69.1} & 75.5 & 85.7 & 82.5\\
    DPL\textsuperscript{$\dagger$}~\cite{zhang2021amortized}  & 99.8 & 61.5 & 77.8 & 84.6  & 80.9\\
    VPT\textsuperscript{$\dagger$}~\cite{jia2022visual}  & \textbf{99.9} & 64.8 & 78.2 & 85.2 & 82.0\\
    MaPLe\textsuperscript{$\dagger$}~\cite{khattak2023maple} & 98.3 & 64.8 & 80.6 & 85.1 & 82.2 \\
    % PromptSRC*~\cite{khattak2023self} & \textbf{99.9} & 67.6 & 77.2 & \textbf{87.6} & 83.1 \\
    SPG~\cite{bai2025soft}  & 99.7 & 64.7 & 80.7 & 84.4 & 82.4\\
    \textcolor{black}{Gallop*~\cite{lafon2024gallop}} & \textcolor{black}{99.8} & \textcolor{black}{68.8} & \textcolor{black}{79.3} & \textcolor{black}{\textbf{87.3}} & \textcolor{black}{\textbf{83.8}}\\
    % Gallop*~\cite{lafon2024gallop}  & 99.8 & 68.8 & 79.3 & \textbf{87.3} & \textbf{83.8}\\
    \textbf{SRE (Ours)}  & 99.0  & 67.8  & \textbf{81.5} & 86.4 & 83.7\\
    \bottomrule
\end{tabular}}
\end{table}
\vspace{-5mm}
\begin{table}[htbp]
\caption{Comparison results between SRE and other CLIP-based methods on OfficeHome under the leave-one-domain-out setting. ``*" denotes our reproduced results, ``$\dagger$" denotes results copied from SPG~\cite{bai2025soft}
  }
  \label{tab:detailed_off}
  \centering
\scalebox{0.9}{\begin{tabular}{l|p{1.cm}<{\centering}|p{1.cm}<{\centering}|p{1.cm}<{\centering}|p{1.cm}<{\centering}|p{0.6cm}<{\centering}}
    \toprule
    {\bf Method} & {\bf Art} & {\bf Clipart}&{\bf Product}&{\bf Real}& {\bf Avg.}\\ \midrule
    % \midrule
    ZS-CLIP~\cite{radford2021learning} & 82.1 & 67.9 & 89.1 & 89.2 & 82.1\\
    DPL\textsuperscript{$\dagger$}~\cite{zhang2021amortized}  & 81.0 & 71.2 & 90.0 & 89.6 & 83.0\\
    VPT\textsuperscript{$\dagger$}~\cite{jia2022visual}  & 80.9 & 72.5 & 89.0 & 90.4 & 83.2\\
    MaPLe\textsuperscript{$\dagger$}~\cite{khattak2023maple} & 81.6 & 72.6 & 90.2 & 89.0 & 83.4 \\
    % PromptSRC*~\cite{khattak2023self} & 84.9 & 70.7 & 92.3 & 92.0 & 85.0 \\
    SPG~\cite{bai2025soft}  & 81.6 & 72.7 & 90.2 & 89.9 & 83.6\\
    % Gallop*~\cite{lafon2024gallop}  & 85.9 & \textbf{73.9} & \textbf{92.7} & 91.5 & 86.0\\
    \textcolor{black}{Gallop*~\cite{lafon2024gallop}} & \textcolor{black}{85.9} & \textcolor{black}{\textbf{73.9}} & \textcolor{black}{\textbf{92.7}} & \textcolor{black}{91.5} & \textcolor{black}{86.0}\\
    \textbf{SRE (Ours)}  & \textbf{86.8}  & 73.0  & 92.4 & \textbf{92.1} & \textbf{86.1}\\
    \bottomrule
    \end{tabular}}
\end{table}
\vspace{-5mm}
\begin{table}[htbp]
\caption{Comparison results between SRE and other CLIP-based methods on TerraIncognita under the leave-one-domain-out setting. ``*" denotes our reproduced results, ``$\dagger$" denotes results copied from SPG~\cite{bai2025soft}
  }
  \label{tab:detailed_terra}
  \centering
\scalebox{0.9}{\begin{tabular}{l|p{1.cm}<{\centering}|p{1.cm}<{\centering}|p{1.cm}<{\centering}|p{1.cm}<{\centering}|p{0.6cm}<{\centering}}
    \toprule
    {\bf Method} & {\bf L100} & {\bf L38}&{\bf L43}&{\bf L46}& {\bf Avg.}\\ \midrule
    % \midrule
    ZS-CLIP~\cite{radford2021learning} & 52.1 & 19.9 & 33.5 & 30.0 & 33.9\\
    DPL\textsuperscript{$\dagger$}~\cite{zhang2021amortized} & 41.6  & 54.3 & 49.0 & 41.6 & 46.6\\
    VPT\textsuperscript{$\dagger$}~\cite{jia2022visual} & 45.5  & 46.8 & 52.8 & 41.8 & 46.7\\
    MaPLe\textsuperscript{$\dagger$}~\cite{khattak2023maple} & 52.4 & 52.4 & 53.0 & 43.1 & 50.2 \\
    % PromptSRC*~\cite{khattak2023self} & 56.5 & 46.3 & 53.7 & 44.3 & 50.2 \\
    SPG~\cite{bai2025soft}  & 49.8 & 51.0 & 49.2 & \textbf{50.7} & 50.2\\
    % Gallop*~\cite{lafon2024gallop}  & 59.3 & 51.8 & 52.6 & 43.3 & 51.7\\
    \textcolor{black}{Gallop*~\cite{lafon2024gallop}} & \textcolor{black}{59.3} & \textcolor{black}{51.8} & \textcolor{black}{52.6} & \textcolor{black}{43.3} & \textcolor{black}{51.7}\\
    \textbf{SRE (Ours)}  & \textbf{73.7}  & \textbf{58.4} & \textbf{61.1} & 49.8 & \textbf{60.8}\\
  \bottomrule
  \end{tabular}}
\end{table}
\vspace{-5mm}
\begin{table}[ht]
\caption{Comparison results between SRE and other CLIP-based methods on DomainNet under the leave-one-domain-out setting. ``*" denotes our reproduced results, ``$\dagger$" denotes results copied from SPG~\cite{bai2025soft}
  }
  \label{tab:detailed_dom}
  \centering
  \scalebox{0.85}{\begin{tabular}{l|p{0.8cm}<{\centering}|p{0.8cm}<{\centering}|p{0.8cm}<{\centering}|p{0.8cm}<{\centering}|p{0.8cm}<{\centering}|p{0.8cm}<{\centering}|p{0.4cm}<{\centering}}
    \toprule
    {\bf Method} & {\bf Clipart} & {\bf Info}&{\bf Paint}&{\bf Quick}&{\bf Real}&{\bf Sketch}& {\bf Avg.}\\ \midrule 
    ZS-CLIP~\cite{radford2021learning} & 71.2 & 48.1 & 66.3 & 14.0 & 83.4 & 63.2 & 57.7\\
    DPL\textsuperscript{$\dagger$}~\cite{zhang2021amortized}  & 72.5 & 50.4 & 68.3 & 15.8 & 83.9 & 66.0 & 59.5\\
    VPT\textsuperscript{$\dagger$}~\cite{jia2022visual}  & 71.0 & 48.5 & 66.2 & 16.3 & 83.6 & 65.2 & 58.5\\
    MaPLe\textsuperscript{$\dagger$}~\cite{khattak2023maple} & 73.1 & 49.9 & 67.8 & 16.6 & 83.5 & 65.9 & 59.5\\
    SPG~\cite{bai2025soft}  & 68.7 & 50.2 & \textbf{73.2} & 16.6 & 83.3 & 68.5 & 60.1\\
    % Gallop*~\cite{lafon2024gallop}  & 76.1 & \textbf{54.2} & 70.9 & 17.3 & \textbf{84.5} & 68.0 & \textbf{61.8}\\
    \textcolor{black}{Gallop*~\cite{lafon2024gallop}} & \textcolor{black}{76.1} & \textcolor{black}{\textbf{54.2}} & \textcolor{black}{70.9} & \textcolor{black}{17.3} & \textcolor{black}{\textbf{84.5}} & \textcolor{black}{68.0} & \textcolor{black}{\textbf{61.8}}\\
    \textbf{SRE (Ours)}  & \textbf{76.2}  & 52.0  & 70.3 & \textbf{18.9} & 84.3 & \textbf{68.9} & \textbf{61.8}\\
  \bottomrule
  \end{tabular}}
\end{table}

% 新增的SDG实验在这里！
\textcolor{black!70!black}{\subsection{Single Domain Generalization}}
\textcolor{black!70!black}{To investigate the effectiveness of our method on single domain generalization, we conduct comparative experiments on the OfficeHome~\cite{venkateswara2017deep} and TerraIncognita~\cite{beery2018recognition} datasets with other CLIP-based methods, including DPL~\cite{zhang2021amortized}, VPT~\cite{jia2022visual} and MaPLe~\cite{khattak2023maple}. To measure the performance of a model on a certain domain, we train the model on this domain, test it on the left three domains, and average the results of the model on the three domains. As shown in Table~\ref{tab:sdg_off} and Table~\ref{tab:sdg_terra}, the proposed SRE achieves the highest average accuracy on both datasets. % and outperforms the second-best method by 6.8\% on TerraIncognita, demonstrating the ability of our method in single domain generalization.
}

\begin{table}[htbp]
    \centering
    \color{black!70!black}
    \caption{Comparison results between SRE and other CLIP-based methods on OfficeHome under single domain generalization setting. ``*" denotes our reproduced results.}
    \scalebox{0.9}{\begin{tabular}{l|p{1.cm}<{\centering}|p{1.cm}<{\centering}|p{1.cm}<{\centering}|p{1.cm}<{\centering}|p{0.6cm}<{\centering}}
    \toprule
    {\bf Method} & {\bf Art} & {\bf Clipart}&{\bf Product}&{\bf Real}& {\bf Avg.}\\ \midrule
    % \midrule
    ZS-CLIP~\cite{radford2021learning} & 82.1 & 86.8 & 79.7 & 79.7 & 82.1\\
    DPL*~\cite{zhang2021amortized}  & \textbf{83.7} & 87.4 & 80.6 & 81.6 & 83.3\\
    VPT*~\cite{jia2022visual}  & 78.1 & 82.1 & 75.1 & 78.9 & 78.5\\
    MaPLe*~\cite{khattak2023maple} & 83.0 & 87.3 & 80.6 & 81.4 & 83.1 \\
    \textbf{SRE (Ours)}  & 83.1  & \textbf{87.8}  & \textbf{82.1} & \textbf{83.5} & \textbf{84.1}\\
    \bottomrule
    \end{tabular}}
    \label{tab:sdg_off}
\end{table}
\begin{table}[htbp]
    \centering
    \color{black!70!black}
    \caption{Comparison results between SRE and other CLIP-based methods on TerraIncognita under single domain generalization setting. ``*" denotes our reproduced results.}
    \scalebox{0.9}{\begin{tabular}{l|p{1.cm}<{\centering}|p{1.cm}<{\centering}|p{1.cm}<{\centering}|p{1.cm}<{\centering}|p{0.6cm}<{\centering}}
    \toprule
    {\bf Method} & {\bf L100} & {\bf L38}&{\bf L43}&{\bf L46}& {\bf Avg.}\\ \midrule
    % \midrule
    ZS-CLIP~\cite{radford2021learning} & 27.8 & 38.5 & 34.0 & 35.2 & 33.9\\
    DPL*~\cite{zhang2021amortized} & 42.4  & 29.8 & 40.9 & 50.3 & 40.9\\
    VPT*~\cite{jia2022visual} & 34.7  & 23.0 & 42.8 & 46.5 & 36.5\\
    MaPLe*~\cite{khattak2023maple} & 41.2 & 33.5 & 44.8 & 46.6 & 41.5 \\
    \textbf{SRE (Ours)}  & \textbf{45.1}  & \textbf{36.0} & \textbf{55.1} & \textbf{52.9} & \textbf{47.3}\\
  \bottomrule
  \end{tabular}}
    \label{tab:sdg_terra}
\end{table}
\textcolor{black}{\subsection{Open-Set Domain Generalization}}
\textcolor{black}{To further explore the potential of our method when the target domain exhibits semantic shifts with source domains, we conduct experiments on OfficeHome and TerraIncognita datasets under the Open-Set Domain Generalization (OSDG) setting. OSDG constructs a scenario where both data distribution shift and semantic shift exist between the target domain and the source domain, and the target domain contains unknown categories that do not belong to the source domain's category space. In this scenario, the model must not only accurately classify known class samples outside the source distribution, but also distinguish samples of unknown classes.}

\textcolor{black}{
To make a fair comparison with other CLIP-based methods, we combine all methods with a common open-set recognition strategy that distinguishes unknown classes by setting a specific threshold for prediction probabilities. Following previous works, we choose three evaluation metrics to verify the performance: the close-set accuracy (Acc), the harmonic mean of known class accuracy and unknown class accuracy (H-score), and the open-set classification rate (OSCR), which plots the true positive rate against the false positive rate by a varying threshold.
}

\textcolor{black}{Table~\ref{tab:osdg1} and Table~\ref{tab:osdg2} show the  results on both OfficeHome and TerraIncognita datasets, which validates the effectiveness of SRE in Open-Set Domain Generalization. SRE not only achieves the highest close-set accuracy (Acc), but also excels in H-score and OSCR, demonstrating its ability to accurately classify known classes while effectively identifying unknown classes. Compared to existing CLIP-based methods, SRE consistently outperforms baselines in handling domain shifts and semantic shifts, making it a robust solution in real-world applications where unknown categories are prevalent.
}

\begin{table*}[htbp]
    \centering
    \color{black}
    \caption{Comparison results between SRE and other CLIP-based methods on OfficeHome under the Open-Set Domain Generalization setting. The ratio of known to unknown classes is 50:15. ``*" denotes our reproduced results.}
    \label{tab:osdg1}
    \scalebox{0.9}{\begin{tabular}{l|ccc|ccc|ccc|ccc|ccc}
        \toprule
        \multirow{2}{*}{\bf Method} & \multicolumn{3}{c|}{\bf Art} & \multicolumn{3}{c|}{\bf Clipart} & \multicolumn{3}{c|}{\bf Product} & \multicolumn{3}{c|}{\bf Real World} & \multicolumn{3}{c}{\bf Avg} \\
        & Acc & H-score & OSCR & Acc & H-score & OSCR &  Acc & H-score & OSCR & Acc & H-score & OSCR &  Acc & H-score & OSCR \\
        \midrule
        ZS-CLIP~\cite{radford2021learning} & 84.1 & 75.9 & 76.6 & 70.3 & 65.9 & 61.8 & 90.0 & 80.2 & 83.5 & 90.6 & 82.2 & 85.4 & 83.7 & 76.1 & 76.8 \\
        DPL*~\cite{zhang2021amortized} & 85.0 & 74.5 & 75.5 & 73.7 & 66.7 & 64.0 & {\bf 92.9} & {\bf 81.9} & {\bf 86.2} & 91.8 & 82.6 & 85.9 & 85.9 & 76.4 & 77.9 \\
        VPT*~\cite{jia2022visual} & 83.8 & 75.3 & 75.8 & 74.1 & 67.1 & 64.8 & 91.4 & 81.1 & 85.0 & 91.3 & 81.7 & 85.6 & 85.1 & 76.3 & 77.8 \\
        MaPLe*~\cite{khattak2023maple} & 84.7 & 75.1 & 76.4 & 73.3 & 66.1 & 63.3 & 92.1 & 81.7 & 85.9 & 92.3 & {\bf 83.4} & 86.9 & 85.6 & 76.6 & 78.1 \\
        SRE (Ours) & {\bf 87.5} & {\bf 77.5} & {\bf 79.7}  & {\bf 75.7} & {\bf 67.2} & {\bf 65.5} & {\bf 92.9} & 81.8 & {\bf 86.2} & {\bf 93.0} & 83.1 & {\bf 87.9} & {\bf 87.3} & {\bf 77.4} & {\bf 79.8}  \\
        \bottomrule
    \end{tabular}}
\end{table*}

\begin{table*}[htbp]
    \centering
    \color{black}
    \caption{Comparison results between SRE and other CLIP-based methods on TerraIncognita under the Open-Set Domain Generalization setting. The ratio of known to unknown classes is 8:2. ``*" denotes our reproduced results.}
    \label{tab:osdg2}
    \scalebox{0.9}{\begin{tabular}{l|ccc|ccc|ccc|ccc|ccc}
        \toprule
        \multirow{2}{*}{\bf Method} & \multicolumn{3}{c|}{\bf Location 100} & \multicolumn{3}{c|}{\bf Location 38} & \multicolumn{3}{c|}{\bf Location 43} & \multicolumn{3}{c|}{\bf Location 46} & \multicolumn{3}{c}{\bf Avg} \\
        & Acc & H-score & OSCR & Acc & H-score & OSCR &  Acc & H-score & OSCR & Acc & H-score & OSCR &  Acc & H-score & OSCR \\
        \midrule
        ZS-CLIP~\cite{radford2021learning} & 44.2 & 44.8 & 36.3 & 26.1 & 29.3 & 19.4 & 38.8 & 40.5 & 31.2 & 32.6 & 28.8 & 19.5 & 34.3 & 34.3 & 25.1 \\
        DPL*~\cite{zhang2021amortized} & 64.4 & 54.1 & 46.7 & 61.0 & 55.5 & 49.3 & 56.9 & 49.3 & 42.4 & 47.6 & 43.6 & 34.5 & 57.5 & 50.6 & 43.3 \\
        VPT*~\cite{jia2022visual} & 61.1 & 48.3 & 40.6 & 55.3 & 53.3 & 45.4 & 56.7 & 49.4 & 41.1 & 50.9 & 46.2 & 37.8 & 56.0 & 49.3 & 41.2 \\
        MaPLe*~\cite{khattak2023maple} & 68.4 & 55.7 & 51.3 & 61.3 & 55.9 & 49.6 & 58.6 & 51.8 & 44.6 & 48.1 & 43.0 & 34.3 & 59.1 & 51.6 & 45.0 \\
        SRE (Ours) & {\bf 75.8} & {\bf 68.3} & {\bf 65.2}  & {\bf 64.5} & {\bf 58.0} & {\bf 53.1} & {\bf 61.3} & {\bf 59.6} & {\bf 53.7} & {\bf 54.9} & {\bf 53.0} & {\bf 45.2} & {\bf 64.1} & {\bf 59.7} & {\bf 54.3}  \\
        \bottomrule
    \end{tabular}}
\end{table*}

\vspace{3pt}
\subsection{Ablation Studies}

\begin{table*}[htbp]
	\caption{Results of different stages on PACS, VLCS, OfficeHome,  Terra, and DomainNet with ViT-B/16 as the backbone. \textcolor{black}{``SR + EMA" denotes replacing the Ensemble stage with the Exponential Moving Average (EMA), which serves as a baseline for the parameter ensembling.}
	}
	\label{tab:ablation2}
	\centering
	\scalebox{0.9}{\begin{tabular}{l|p{1.2cm}<{\centering}p{1.2cm}<{\centering}p{1.2cm}<{\centering}|p{1.2cm}<{\centering}p{1.2cm}<{\centering}p{1.2cm}<{\centering}p{1.2cm}<{\centering}p{1.2cm}<{\centering}p{0.6cm}<{\centering}}
			\toprule
			{\bf Method} & {\bf Refocus} & {\bf Simulate}&{\bf Ensemble}&{\bf PACS}&{\bf VLCS}&{\bf OfficeHome} &{\bf Terra}&{\bf DomainNet}&{\bf Avg.}\\
			\midrule
			ZS-CLIP~\cite{radford2021learning}  &  &  &  & 96.1 &82.5& 82.1 & 33.9 &57.5& 70.4\\
			AR-CLIP & \checkmark &  &  & 95.9 &82.8& 83.6 & 54.3 &59.7& 75.3\\
			SR & \checkmark & \checkmark &  & 96.3 &83.3& 84.3 & 59.8 &61.2& 77.0\\
			\textcolor{black}{SR + EMA} & \checkmark & \checkmark &  & 96.1 &83.5& 84.6 & 60.0 &61.6& 77.1\\
			\midrule
			\textbf{SRE (Ours)}  & \checkmark & \checkmark & \checkmark & \textbf{97.0} &\textbf{83.7}& \textbf{86.1} & \textbf{60.8} &\textbf{61.8}& \textbf{77.9}\\
			\bottomrule
	\end{tabular}}
\end{table*}
\vspace{-3pt}
\paragraph{Results of different stages} To evaluate the effectiveness of each stage of our method, %we conduct ablation experiments on PACS, OfficeHome, and Terra. 
We compare our method with the following variants: 
(1) ZS-CLIP, where the original CLIP is used to perform zero-shot classification; 
(2) AR-CLIP, where only the attention-refocuser is incorporated into ZS-CLIP for fine-tuning; 
(3) SR, where the \emph{Ensemble} stage is removed from the SRE scheme; 
(4) SR + EMA, where an exponential moving average strategy is added to SR to average all weights instead of selecting and combining parameters as done in the \emph{Ensemble} stage.

Table~\ref{tab:ablation2} shows the results of different modules on the five datasets. 
%, where we use ``RE", ``SI" and ``EN" to denote the \emph{Refocus}, \emph{Simulate}, and \emph{Ensemble} stages respectively.
It is interesting to observe that: (1) AR-CLIP outperforms ZS-CLIP, which suggests that attention refocusing improves the generalization capability of CLIP; (2) SR outperforms AR-CLIP, which indicates that the simulated target domains and attention maps aligning boost the generalization of attention refocusing; (3) SR + EMA performs worse than  SRE and similar to SR, which indicates that the performance improvement of the \emph{Ensemble} stage is not benefited from simple weight average, but from adaptive selection to the robust parameters.

In addition, we explicitly compare our method with TOAST~\cite{shi2023toast} on five datasets and report the results in Table~\ref{tab:ablation3}. We observe that directly applying TOAST to fine-tune the CLIP image encoder can only achieve marginal performance gains. Our combination of the attention-refocuser and zero-shot classification, denoted as AR-CLIP, alleviates over-fitting to some extent. Most importantly, the proposed SRE greatly improves DG performance.
\begin{table}[htbp]
	\caption{Comparison results between TOAST~\cite{shi2023toast} and our method on PACS, VLCS, OfficeHome,  Terra, and DomainNet with ViT-B/16 as the backbone.}
	\label{tab:ablation3}
	\centering
	\scalebox{0.95}{\begin{tabular}{l|p{0.8cm}<{\centering}p{0.8cm}<{\centering}p{0.8cm}<{\centering}p{0.8cm}<{\centering}p{0.8cm}<{\centering}p{0.6cm}<{\centering}}
			\toprule
			{\bf Method} & {\bf PACS}&{\bf VLCS}&{\bf Off.H.} &{\bf Terra}&{\bf Dom.N.}&{\bf Avg.}\\
			\midrule
			% ZS-CLIP~\cite{radford2021learning}  & 96.1 &82.5& 82.1 & 33.9 &57.5& 70.4\\
                TOAST~\cite{shi2023toast} & 95.9 &82.1& 83.8 & 44.0 &58.6& 72.9\\
			AR-CLIP & 95.9 &82.8& 83.6 & 54.3 &59.7& 75.3\\
			\textbf{SRE (Ours)}  & \textbf{97.0} &\textbf{83.7}& \textbf{86.1} & \textbf{60.8} &\textbf{61.8}& \textbf{77.9}\\
			\bottomrule
	\end{tabular}}
\end{table}

\vspace{-3pt}
\textcolor{black}{\paragraph{Results of different losses}
We conduct ablation studies on different losses in Eq.~\ref{equ:loss_all}. 
As shown in Table~\ref{tab:ablation_loss}, the combination of $\mathcal{L}_{cls}$ and $\mathcal{L}_{align}$ achieves significantly better performance than either loss alone, demonstrating that the design in the \emph{Simulate} stage can better align the source domain and simulation target domain. The introduction of $\mathcal{L}_{var}$ further improves the robustness, especially in challenging domains like Terra (+2.0\% with $\mathcal{L}_{cls} + \mathcal{L}_{var}$). Our full model integrating all three losses achieves the best performance (75.8\% average, 59.8\% on Terra), verifying that $\mathcal{L}_{var}$ effectively stabilizes the joint optimization process while preserving alignment benefits.}

\begin{table}[htbp]
	\centering
        \color{black}
        \caption{Results of different loss on VLCS, OfficeHome,  Terra with ViT-B/16 as the backbone.}
	\scalebox{0.88}{\begin{tabular}{l|p{0.7cm}<{\centering}p{0.7cm}<{\centering}p{0.7cm}<{\centering}|p{0.6cm}<{\centering}p{0.6cm}<{\centering}p{0.6cm}<{\centering}p{0.6cm}<{\centering}}
			\toprule
			{\bf Method} & {\bf $\mathcal{L}_{cls}$} & {\bf $\mathcal{L}_{align}$}&{\bf $\mathcal{L}_{var}$}&{\bf VLCS}&{\bf Off.H.} &{\bf Terra}&{\bf Avg.}\\
			\midrule
			ZS-CLIP~\cite{radford2021learning}  &  &  &  &82.5& 82.1 & 33.9 & 70.4\\
			+ $\mathcal{L}_{cls}$ & \checkmark &  &  &82.8& 83.6 & 54.3 & 73.6\\
			+ $\mathcal{L}_{cls}, \mathcal{L}_{var}$ & \checkmark &  & \checkmark &83.5 & 83.2 & 55.3 & 74.0\\
			+ $\mathcal{L}_{align}$ &  & \checkmark &  &83.0& 82.1 & 54.1 & 73.1\\
                + $\mathcal{L}_{align}, \mathcal{L}_{var}$ &  & \checkmark & \checkmark &83.4& 82.2 & 54.9 & 73.5\\
                + $\mathcal{L}_{cls}, \mathcal{L}_{align}$ & \checkmark & \checkmark &  &\textbf{83.7}& 85.9 & 60.1 & 76.5\\
			\midrule
			\textbf{SR (Ours)}  & \checkmark & \checkmark & \checkmark & \textbf{83.7} & \textbf{86.1} & \textbf{60.8} & \textbf{76.9}\\
			\bottomrule
	\end{tabular}}
    \label{tab:ablation_loss}
\end{table}

\vspace{-3pt}
\textcolor{black}{\paragraph{Analysis on the \emph{Simulate} stage}
We analyze the impact of using different data augmentation strategies in the \emph{Simulate} stage. 
We use the combination of used augmentation (ColorJitter, GaussianBlur, and GrayScale) as a baseline method.
We explore two types of variants: (1) Incremental variants (adding Polarize, Equalize, Solarize, or their combination to the baseline); (2) Alternative variants (replacing the entire baseline with Fourier Transform~\cite{xu2021fourier} or Random Convolution~\cite{xu2020robust}). 
Table~\ref{tab:aug_ablation} presents the results on the VLCS, OfficeHome, and TerraIncognita datasets, where the baseline method achieves stable cross-domain performance. Incremental variants slightly improve the accuracy on VLCS but degrade the performance on TerraIncognita, suggesting that excessive enhancement may hurt feature extraction. Alternative variants exhibit task-dependent effectiveness: Fourier transform performs well on OfficeHome but poorly on TerraIncognita, while random convolution reduces performance across all datasets. These results show that adding more augmentation strategies does not bring a larger overall performance improvement. Our choice reflects a systematic trade-off between simplicity and performance.}

\begin{figure}[htbp]
	\centering
		\includegraphics[width=0.9\linewidth]{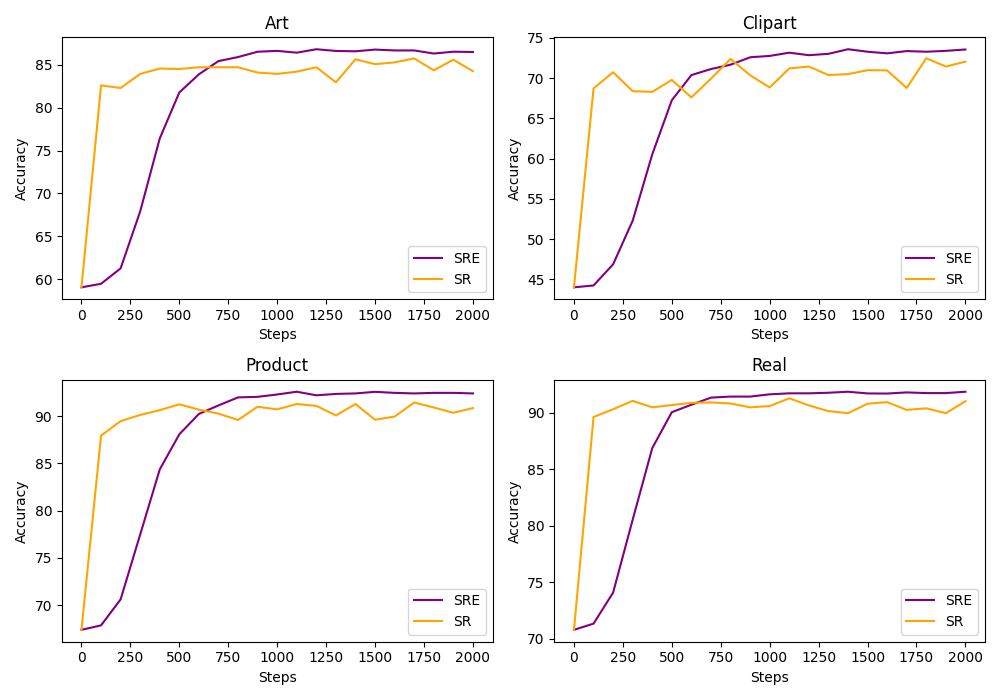}
		\caption{
			Comparison of the test accuracy between SRE and its variant SR on the four target domains of the OfficeHome dataset.
		}
		\label{fig:ablation_E}
\end{figure}

\textcolor{black}{To further investigate the generalization ability of our model on unseen simulated target domains using the three augmentation strategies, we apply other augmentations to the target domains to emulate a broader spectrum of distribution shifts. 
%we evaluate our model on unseen simulated target domains by applying other augmentations to the target domains, emulating a broader spectrum of distribution shifts. 
As shown in Table~\ref{tab:aug_ablation2}, we add augmentation strategies, including Equalize, Polarize, Solarize, and their combination (``ALL"), as well as Fourier Transform and Random Convolution, to the test set of VLCS, OfficeHome, and TerraIncognita. The results demonstrate that our method (SRE) consistently outperforms all compared methods under all simulated shifts. These findings underscore that  SRE exhibits better generalization across diverse domain shifts, including those far beyond the original augmentation (e.g., Fourier transform and random convolution), suggesting its potential applicability to real-world domain shifts.}
\setcounter{table}{12}
\begin{table*}[!htbp]
\centering
\color{black}
\caption{Results of our method on VLCS, OfficeHome, and TerraIncognita when adding different data augmentation strategies on target domain to simulate various domain shifts. ``All" denotes combining Polarize, Equalize, and Solarize together.}
\label{tab:aug_ablation2}
\scalebox{0.8}{\begin{tabular}{l|cccccccccccc|cccccc}
\toprule
 \multirow{2}{*}{\textbf{Method}} & \multicolumn{3}{c}{\textbf{Equalize}} & \multicolumn{3}{c}{\textbf{Polarize}} & \multicolumn{3}{c}{\textbf{Solarize}} & \multicolumn{3}{c|}{\textbf{ALL}} & \multicolumn{3}{c}{\textbf{Fourier}} & \multicolumn{3}{c}{\textbf{RandConv}} \\
\cmidrule(lr){2-4} \cmidrule(lr){5-7} \cmidrule(lr){8-10} \cmidrule(lr){11-13} \cmidrule(lr){14-16} \cmidrule(lr){17-19}
 & VLCS & Off.H. & Terra & VLCS & Off.H. & Terra & VLCS & Off.H. & Terra & VLCS & Off.H. & Terra & VLCS & Off.H. & Terra & VLCS & Off.H. & Terra \\
\midrule
ZS-CLIP~\cite{radford2021learning} & 82.2 & 80.2 & 34.5 & 82.5 & 81.0 & 27.5 & 81.7 & 79.2 & 33.8 & 81.4 & 76.5 & 27.5 & 82.6 & 75.9 & 31.4 & 80.1 & 72.0 & 21.8 \\
VPT*~\cite{jia2022visual} & 82.3 & 81.9 & 44.8 & 82.2 & 82.6 & 38.2 & 81.7 & 80.9 & 44.6 & 81.7 & 78.5 & 35.7 & 82.7 & 78.1 & 40.2 & 80.0 & 75.7 & 26.3 \\
MaPLe*~\cite{khattak2023maple} & 82.2 & 82.3 & 46.3 & 82.0 & 83.2 & 39.9 & 82.2 & 81.3 & 45.9 & 82.0 & 78.8 & 37.5 & 82.8 & 78.7 & 42.9 & 81.5 & 75.7 & 29.2 \\
\textbf{SRE (Ours)} & \textbf{83.6} & \textbf{84.6} & \textbf{59.7} & \textbf{83.7} & \textbf{85.2} & \textbf{48.6} & \textbf{83.4} & \textbf{84.0} & \textbf{59.1} & \textbf{83.3} & \textbf{81.8} & \textbf{46.8} & \textbf{83.3} & \textbf{81.7} & \textbf{56.7} & \textbf{81.8} & \textbf{78.9} & \textbf{42.4} \\
\bottomrule
\end{tabular}}
\end{table*}
\setcounter{table}{11}
\begin{table}[htbp]
\centering
\color{black}
\caption{Results of our method on VLCS, OfficeHome, and TerraIncognita when using different data augmentation strategies. ``CJ", ``GB", and ``GS" denote ColorJitter, GaussianBlur, and GrayScale, respectively. ``All" denotes combining Polarize, Equalize, and Solarize together. The row in gray denotes the augmentation strategies used in SRE.}
\label{tab:aug_ablation}
\scalebox{0.9}{\begin{tabular}{l|l|ccc}
\toprule
\textbf{Category} & \textbf{Augmentation Variant} & \textbf{VLCS} & \textbf{Off.H.} & \textbf{Terra}\\
\midrule
ZS-CLIP~\cite{radford2021learning} & None & 82.5 & 82.1 & 33.9 \\ 
\midrule
\multirow{5}{*}{Incremental} 
 & \cellcolor{gray!25}Base (CJ+GB+GS) & \cellcolor{gray!25}83.7 & \cellcolor{gray!25}\textbf{86.1} & \cellcolor{gray!25}\textbf{60.8} \\
 & {Base + Polarize} & \textbf{83.9} & 86.0 & 60.0 \\ 
 & {Base + Equalize} & \textbf{83.9} & 86.0 & 59.5 \\
 & {Base + Solarize} & 83.6 & \textbf{86.1} & 59.7 \\
 & {Base + All} & 83.8 & 86.0 & 59.4 \\
\midrule
\multirow{2}{*}{Alternative} 
 & {Fourier Transform} & {83.7} & 85.7 & 58.3 \\
 & {Random Convolution} & 83.7 & \textbf{86.1} & 58.6 \\
\bottomrule
\end{tabular}}
\end{table}
\setcounter{table}{13}
\paragraph{Ablation on the \emph{Ensemble} stage}
We conducted additional ablation studies on the four domains of the OfficeHome dataset to evaluate the effectiveness of the \emph{Ensemble} stage. 
%As illustrated in 
Figure~\ref{fig:ablation_E} shows the test accuracy of SRE and SR (without the \emph{Ensemble} stage) at intervals of 100 training steps. The results show that SRE demonstrates better generalization capability compared to SR and significantly stabilizes the training process. In the early training stage, SRE performs worse than SR because the ensembled parameters are initialized to zero. This design ensures that the ensembled parameters select the weights of models once the training process is more stable.

\begin{figure}[htbp]
	\centering
		\centering
		\includegraphics[width=0.85\linewidth]{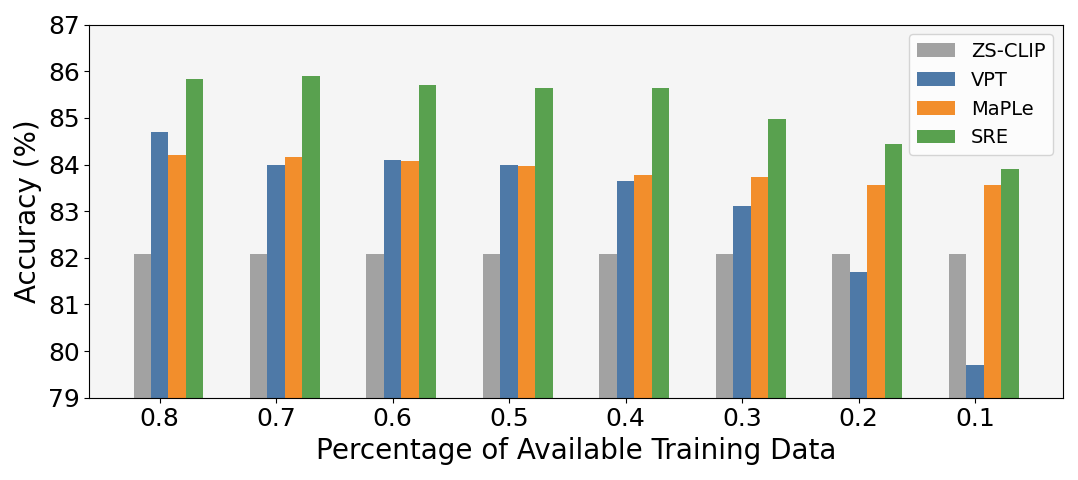}
		\caption{
			Results of percentages of available training data on OfficeHome. 
		}
		\label{fig:holdout}
\end{figure}

\paragraph{Results of different data scales} To evaluate our method in the scenario of insufficient training data, we conduct experiments on the OfficeHome dataset with different training data scales and record the average accuracy across four domains. We gradually decrease the percentage of available training data from 80\% to 10\%. We compare our method with ZS-CLIP~\cite{radford2021learning}, VPT~\cite{jia2022visual} and MaPLe~\cite{khattak2023maple}, and report the results in Figure~\ref{fig:holdout}. 
From the results, we notice that our method outperforms VPT, MaPLE, and ZS-CLIP in all percentages of available training data. Notably, when only 10\% of the data is available for training, our method still achieves performance gains compared to ZS-CLIP, while VPT suffers from overfitting and performs worse generalization ability than ZS-CLIP.
The results verify the capability of our method on insufficient training data.
% The results verify the capability of our method to overcome the problem of overfitting caused by insufficient training data.
% \FloatBarrier

% \subsubsection{Sensitivity to hyperparameters}

\begin{figure}[htbp]
		\centering
		\includegraphics[width=0.6\linewidth]{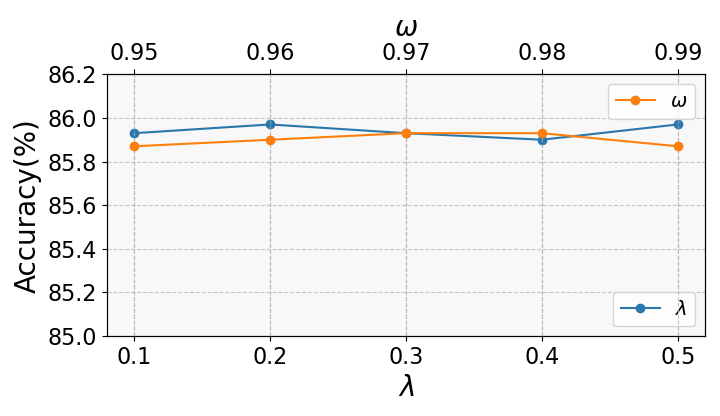}
		% \adjustbox{trim=0 0 0 0,clip}{\includegraphics[width=0.7\linewidth]{img/update_ratio1.png}}
		% \adjustbox{trim=0 0 0 0,clip}{\includegraphics[width=0.7\linewidth]{img/update_ratio2.png}}
		\caption{
			Results of different weights $\lambda$ of the variance loss and different update ratios $\omega$ for ensembling parameter $\theta_a$ on OfficeHome.}
		\label{fig:sensitivity}
\end{figure}

\paragraph{Sensitivity to hyperparameters} 
% Our method introduces two hyperparameters, where $\lambda$ represents the weight of the variance loss $\mathcal{L}_{var}$, and $\omega$ denotes the update ratio of the ensembled parameter $\theta_a$ and the threshold $\hat{s}$. 
We investigate how sensitive our method is to the weight $\lambda$ of the variance loss and the update ratio $\omega$ by changing their values and reporting the corresponding results.  
Figure~\ref{fig:sensitivity} shows the average accuracy over three runs on OfficeHome. The results indicate that our method is insensitive to both $\lambda$ and $\omega$.

% \begin{figure}[tb]
	%   \centering
	%   \subfigure[Visualization of DV-CLIP compared with previous methods]{\includegraphics[width=0.47\textwidth]{img/grad_cam.png}}
	%   \quad\quad\quad
	%   % \includegraphics[width=0.42\textwidth]{img/grad_cam.png}
	%     \subfigure[Visualization of the effectiveness of the two modules in DV-CLIP]{\includegraphics[width=0.3\textwidth]{img/grad_refine.png}}
	
	%   \caption{
		%   Visualization results of heatmaps generated by Grad-CAM~\cite{selvaraju2017grad} for our method and comparative methods including DPL, VPT and the variant DV-CLIP-aug.\wzy{add images in Terra}.
		%   % We present the focus of our method in the form of heatmaps generated by gradCAM~\cite{selvaraju2017grad} and compare our visualization results with prompt tuning methods DPL and VPT.
		%   }
	%   \label{fig:gradcam}
	% \end{figure}

\subsection{Visualization Study}
\begin{figure}[htbp]
	\centering
		% \centering
		\includegraphics[width=0.78\linewidth]{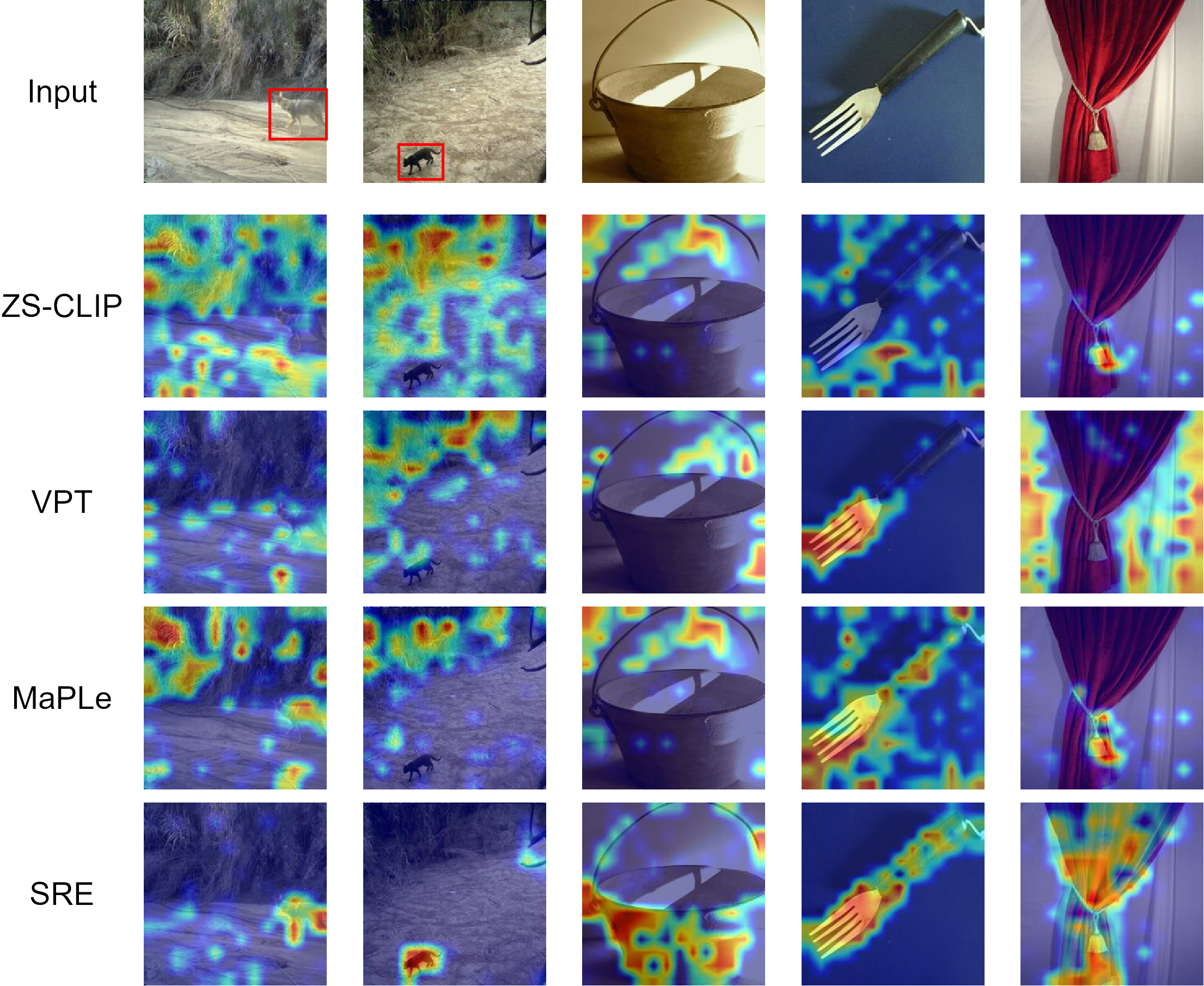}
		\caption{Attention visualization of SRE and several existing methods.}
		\label{fig:gram_comp}
\end{figure}

To further demonstrate the capability of refocusing attention of our method, we visualize CLIP's attention using  Grad-CAM~\cite{selvaraju2017grad} in the form of heatmaps. As shown in Figure~\ref{fig:gram_comp}, our method focuses better on task-relevant regions while paying less attention to task-irrelevant regions, especially on images with noisy backgrounds, which verifies the effectiveness of our method to refocus the attention in CLIP.

\begin{figure}[htbp]
		\centering
		\includegraphics[width=0.77\linewidth]{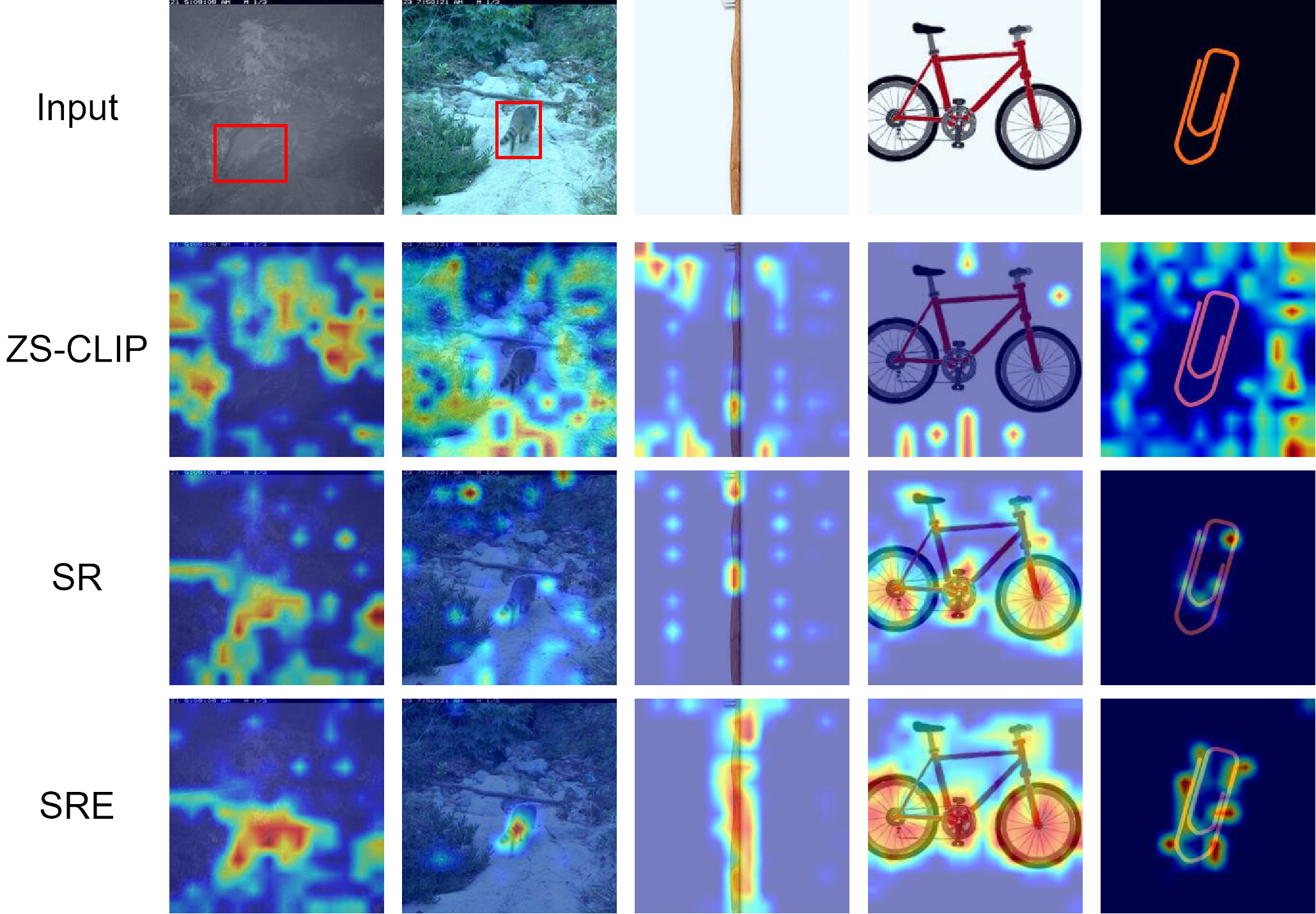}
		\caption{Attention visualization of SRE and several variants.}
		\label{fig:gradcam}
\end{figure}

To further demonstrate the effectiveness of the proposed \emph{Simulate, Refocus and Ensemble} scheme, we present visualization results of attention maps of our method and several variants in  Figure~\ref{fig:gradcam}. 
%We also conduct visualization experiments to verify the effectiveness of the two modules in the proposed DV-CLIP. As demonstrated in Figure~\ref{fig:gradcam}, the dual-view attention refocusing module improves the focus of CLIP on task-relevant regions according to the comparison between DV-CLIP-aug and ZS-CLIP. 
% Additionally, the DV-CLIP achieves the best focus on stable task-relevant regions and captures the most visual details. 
We observe that our method performs better than SR in focusing on task-relevant regions and capturing visual details of both easy and difficult images, which indicates the effectiveness of the \emph{Ensemble} stage. Moreover, SR performs better than ZS-CLIP, which suggests that the \emph{Simulate} and the \emph{Refocus} stages improve the focus of CLIP on task-relevant regions.
\vspace{-2mm}
%The comparison between DV-CLIP and DV-CLIP-aug verifies the effectiveness of the dual-view parameter ensembling module, as DV-CLIP presents the best focus on task-relevant regions and captures the most visual details on both easy and difficult samples.

\subsection{Results on  the NICO Challenge}
%In addition to the five most commonly used datasets, 
We also conduct experiments on the  NICO++\cite{zhang2023nico++} dataset, as shown in Table~\ref{tab:nico}.
Unlike traditional DG datasets, NICO++ focuses more on significant distribution shifts in real-world scenarios and recommends evaluating the model's generalization performance across multiple target domains.
% 需要加一段描述来说明NICO与其他数据集的不同
Environmental changes primarily cause the distribution shifts across different domains in NICO++. The NICO++ dataset consists of six domains: Autumn, Rock, Dim, Grass, Outdoor, and Water. Following the guidelines in \cite{zhang2023nico++}, we categorize these six domains into three groups based on environmental similarity: Autumn-Rock, Dim-Grass, and Outdoor-Water. We train the model on two groups while leaving the third group out for testing, which is analogous to the leave-one-domain-out setting. For comparison, we reproduce three methods: DPL\cite{zhang2021amortized}, VPT~\cite{jia2022visual} and MaPLe~\cite{khattak2023maple}. Additionally, ZS-CLIP~\cite{radford2021learning} represents the zero-shot CLIP approach, which uses the template ``a photo of a [class]" for text prompts.

\begin{table}[htbp]
\caption{Comparison results between SRE and other CLIP-based methods on NICO++ for multi-source domain generalization. ``*" denotes our reproduced results.
}
\label{tab:nico} 
\centering
% \begin{tabular}{lp{1.7cm}<{\centering}p{1.7cm}<{\centering}p{1.7cm}<{\centering}p{1.7cm}<{\centering}p{1.7cm}<{\centering}p{1.7cm}<{\centering}}
	\scalebox{0.85}{\begin{tabular}{l|p{0.8cm}<{\centering}p{0.8cm}<{\centering}|p{0.8cm}<{\centering}p{0.8cm}<{\centering}|p{0.8cm}<{\centering}p{0.8cm}<{\centering}|p{0.4cm}<{\centering}}
			\toprule
			{\bf Method} & {\bf Autumn} & {\bf Rock}&{\bf Dim}&{\bf Grass}&{\bf Outdoor}&{\bf Water}&{\bf Avg.}\\
			\midrule
			ZS-CLIP~\cite{radford2021learning} & 89.0 & 89.8 & 87.9 & 90.4 & 85.8 & 82.5 & 87.6\\            
            DPL*~\cite{zhang2021amortized} & 91.1 & 92.0 & \textbf{90.6} & 92.6 & 90.5 & 85.2 & 90.3\\
            VPT*~\cite{jia2022visual} & 89.6 & 91.9 & 89.0 & 92.6 & 90.1 & 85.6 & 89.8\\
            MaPLe*~\cite{khattak2023maple} & 90.7 & 92.2 & 89.8 & 92.5 & 90.4 & 85.8 & 90.2\\
            % PromptSRC*~\cite{khattak2023self} & 91.1 & 92.1 & 89.8 & 92.8 & \textbf{90.8} & 85.9 & 90.4\\
            % MaPLe~\cite{khattak2023maple} & 89.0 & 89.8 & 87.9 & 90.4 & 85.8 & 82.5 & 87.6\\
            % CLIP-LoRA*~\cite{zanella2024low} & 91.1 & 92.5 & 90.2 & 93.0 & 90.7 & 86.4 & 90.6\\
            \textbf{SRE (Ours)} & \textbf{91.4} & \textbf{92.3} & 90.4 & \textbf{93.2} & \textbf{90.8} & \textbf{86.4} & \textbf{90.8}\\
			\bottomrule
	\end{tabular}}
\end{table}
\vspace{-5mm}
\subsection{Complexity Analysis}

We compare the complexity of SRE with other CLIP-based methods on PACS (7 classes) and OfficeHome (65 classes) under the same mini-batch size of 16 for training. As shown in Table~\ref{tab:complexity}, while SRE requires higher memory usage and slightly longer latency during training compared to methods like VPT~\cite{jia2022visual} and MaPLe~\cite{khattak2023maple}, its resource requirements are still lower than DPL~\cite{zhang2021amortized}. These additional resources are essential for enabling SRE to focus on domain-invariant regions, leading to superior generalization performance.
Furthermore, compared to methods like DPL~\cite{zhang2021amortized} and SPG~\cite{bai2025soft}, where the computational cost increases with the number of categories during training, our method achieves a better trade-off between generalization and computational cost.
During testing, SRE demonstrates a balanced profile by requiring less memory to methods such as DPL~\cite{zhang2021amortized}, while maintaining a competitive latency of 3.1 ms/img. 

\begin{table}[htbp]
    \centering
    \caption{Model complexity comparison between different CLIP-based methods using ViT-16/B. %(DPL~\cite{zhang2021amortized}, VPT~\cite{jia2022visual}, MAPLE~\cite{khattak2023maple}, PromptSRC~\cite{khattak2023self}, CLIP-LoRA~\cite{zanella2024low})
    % \wzy{lack: ZS-CLIP}
    }
    \begin{tabular}{l|c|c|c|c}
    \toprule
         \multirow{2}{*}{\textbf{Method}} & \multicolumn{2}{c|}{\textbf{Train}} & \multicolumn{2}{c}{\textbf{Test}} \\ \cline{2-5}
         & Memory & Latency & Memory & Latency \\
         \hline
         \multicolumn{5}{c}{PACS (7 classes)} \\ 
         \hline
         ZS-CLIP~\cite{radford2021learning} & 0 G & 0 ms/img & 2.1 G& 1.5 ms/img\\
         DPL~\cite{zhang2021amortized} & 13.4 G & 3.6 ms/img & 4.1 G& 3.3 ms/img\\
         VPT~\cite{jia2022visual} & 5.2 G& 1.6 ms/img & 2.7 G& 1.5 ms/img\\
         MaPLe~\cite{khattak2023maple} & 5.5 G& 1.8 ms/img & 2.9 G& 1.5 ms/img \\
         % PromptSRC~\cite{khattak2023self} & 7.9 G & 3.1 ms/img & 4.3 G& 1.6 ms/img \\
         SPG~\cite{bai2025soft} & 5.1 G& 16.7 ms/img & 2.1 G& 4.4 ms/img \\
         \textbf{SRE (Ours)} & 11.7 G& 5.2 ms/img & 3.5 G& 3.1 ms/img\\
         % \bottomrule
         \hline
         \multicolumn{5}{c}{OfficeHome (65 classes)} \\ 
         \hline
         ZS-CLIP~\cite{radford2021learning} & 0 G & 0 ms/img & 2.1 G& 1.5 ms/img\\
         DPL~\cite{zhang2021amortized} & 80.6 G & 26.4 ms/img & 15.9 G& 17.5 ms/img\\
         VPT~\cite{jia2022visual} & 5.3 G& 1.6 ms/img & 2.7 G& 1.5 ms/img\\
         MaPLe~\cite{khattak2023maple} & 6.4 G& 1.9 ms/img & 2.9 G& 1.7 ms/img \\
         % PromptSRC~\cite{khattak2023self} & 7.9 G & 3.1 ms/img & 4.3 G& 1.6 ms/img \\
         SPG~\cite{bai2025soft} & 5.1 G& 87.5 ms/img & 2.1 G& 5.1 ms/img \\
         \textbf{SRE (Ours)} & 11.7 G& 5.2 ms/img & 3.5 G& 3.1 ms/img\\
         \bottomrule
    \end{tabular}
    \label{tab:complexity}
\end{table}

\begin{table}[htbp]
    \centering
    \color{black}
    \caption{Model parameters comparison between different CLIP-based methods using ViT-16/B.}
    \begin{tabular}{l|c|c}
    \toprule
         \textbf{Method} & \textbf{Total Parameters} & \textbf{Trained Parameters} \\
         \midrule
         ZS-CLIP~\cite{radford2021learning} & 149.6 M & 0.0 M \\
         DPL~\cite{zhang2021amortized} & 159.6 M & 10.0 M \\
         VPT~\cite{jia2022visual} & 149.7 M & 0.1 M \\
         MaPLe~\cite{khattak2023maple} & 154.4 M & 4.8 M  \\
         SPG~\cite{bai2025soft} & 154.3 M & 4.7 M\\
         \textbf{SRE (Ours)} & 164.4 M & 14.7 M\\
         \bottomrule
    \end{tabular}
    \label{tab:params}
\end{table}

\textcolor{black}{We also compare the total parameters and trained parameters of SRE with other CLIP-based methods. As shown in Table~\ref{tab:params}, SRE has 164.4M parameters, slightly higher than methods like DPL (159.6M) and MaPLe (154.4M). This is due to the introduced attention-refocuser, which refocuses the attention of CLIP on domain-invariant features.
SRE trains 14.7M parameters, more than prompt-based methods (e.g., VPT: 0.1M, MaPLe: 4.8M). This reflects our design philosophy: instead of limiting updates to prompts alone, we introduce an extra network to adaptively adjust the attention of CLIP. While this increases trainable parameters, it ensures richer generalization ability, as evidenced by the performance gains in Tables~\ref{tab:all_result}-\ref{tab:detailed_dom}.
Furthermore, SRE remains far more parameter-efficient than full fine-tuning approaches (e.g., 149.6M trained parameters for full tuning).}

% We compare the complexity of SRE with other CLIP-based methods on PACS with the same batch size of 16. As shown in Table~\ref{tab:complexity}, although our method requires more memory (11.7G) and a longer time (15.8 ms/img) during training, these additional resources are crucial for focusing task-relevant regions and achieving stronger model generalization. During testing, SRE requires comparable memory cost compared with other methods, while achieving the best generalization performance. 

%Although SRE incurs slightly higher complexity during training, these costs are justified by the improvements in model generalization.
%\textcolor{black}{\section{Limitation}
%}

\vspace{-2mm}
\section{Conclusion and Limitation}
We present an attention-refocusing scheme, named \emph{Simulate, Refocus and Ensemble (SRE)}, for domain generalization, which can adaptively refocus CLIP's attention on task-relevant regions for unseen domains and effectively reduce training instability. 
The \emph{Simulate} stage can well mimic diverse domain shifts, the \emph{Refocus} stage learns to reduce domain shifts by aligning the attention maps of the source data and simulated target data in CLIP, and the \emph{Ensemble} stage can effectively enhance the efficiency of attention refocusing by selecting insensitive parameters.
The above designs have been validated via extensive experiments across multiple domain generalization datasets. 
%SRE achieves competitive performance compared to state-of-the-art methods, verifying the capability of SRE to refocus CLIP's attention on critical visual details.  

\textcolor{black}{Although SRE has achieved promising results, it has  larger model complexity and higher computational cost compared to conventional prompt-tuning methods. %, reflected in larger memory usage and longer time consumption during training. 
This may hinder its deployment in resource-constrained environments. To mitigate this issue, future work will explore optimization strategies such as memory optimization, model pruning, and lightweight architecture design to improve efficiency without significantly compromising performance. In addition, we will investigate techniques like knowledge distillation or quantization to further reduce computational overhead.}
% Then we propose dual-view CLIP to tackle this challenge while addressing the problem of misfitting. 
% In the dual module, we apply attention refocusing to fine-tune CLIP while weakly and strongly augment an image to increase diversity. In the second module, we refine the representation of focus of CLIP and ensemble weights that exhibit consistent focus on two views of images. 

\section*{Acknowledgment}
This work was supported by the Shenzhen Science and Technology Program under Grant No. JCYJ20241202130548062, the Natural Science Foundation of Shenzhen under Grant No. JCYJ20230807142703006, and the Key Research Platforms and Projects of the Guangdong Provincial Department of Education under Grant No.2023ZDZX1034.

\bibliographystyle{IEEEtran}
\bibliography{tmm_ref}

\end{document}